\documentclass[final,1p,times]{elsarticle}

\usepackage{amssymb}
\usepackage{amsmath}
\usepackage{booktabs}
\usepackage{comment}
\usepackage{makecell}

\usepackage{lineno}
\usepackage{tabularx}
\usepackage{graphicx}
\usepackage{enumitem}

\journal{Information Fusion}

\begin{document}

\begin{frontmatter}



\title{A Design Framework for operationalizing Trustworthy Artificial Intelligence in Healthcare: Requirements, Tradeoffs and Challenges for its Clinical Adoption}


\author[label1,label2]{Pedro A. Moreno-Sánchez}\author[label2,label3]{Javier Del Ser}
 \author[label1]{Mark van Gils}\author[label1,label4]{Jussi Hernesniemi}


\affiliation[label1]{organization={Faculty of Medicine and Health Technology, Tampere University},
            city={Tampere},
            postcode={33100}, 
            country={Finland}}

\affiliation[label2]{organization={TECNALIA, Basque Research \& Technology Alliance (BRTA)},
            city={Derio},
            postcode={48160}, 
            country={Spain}}
            
\affiliation[label3]{organization={Department of Mathematics, University of the Basque Country (UPV/EHU)},
            city={Leioa},
            postcode={48940}, 
            country={Spain}}
\affiliation[label4]{organization={Tampere Heart Hospital, Tampere University Hospital},
            city={Tampere},
            postcode={33520}, 
            country={Finland}}

\begin{abstract}
The healthcare sector stands as one of the most promising fields where Artificial Intelligence (AI) can deliver groundbreaking advances in the diagnosis and prognosis of diseases and patient care. Over the past decade, the increase of electronically recorded medical data and advancement in computing capability have paved the way for new data-driven solutions that by using, amongst others, medical images, omics, biosignals, and electronic health record data are close to the capabilities of medical experts in detecting and predicting diseases. Despite their potential and revolutionary impact, the deployment of AI solutions in the healthcare sector faces significant barriers, more related to other issues, such as human-technology acceptance, ethics, and regulatory considerations, than technological constraints. To facilitate clinical adoption, these data-driven solutions must adhere to the Trustworthy AI (TAI) paradigm, which addresses critical aspects such as human agency and oversight, algorithmic robustness, privacy and data governance, transparency, bias and discrimination avoidance, and accountability. However, the inherent complexities of the healthcare sector, as reflected in its diverse processes (including screening, diagnosis, prognosis, and treatment) and the variety of stakeholders involved (clinicians, patients, healthcare providers, and regulators), make the adoption of TAI requirements challenging. This work contributes to the field of TAI in healthcare with a design framework that allows developers of  medical AI systems to focus on the various requirements and characteristics that these solutions must meet for the healthcare stakeholders to consider these systems trustworthy. Thus, for each stakeholder identified across various healthcare processes, we propose a disease-agnostic collection of requirements that medical AI systems should incorporate to adhere to the principles of TAI. To make the framework operational, we specify which aspects should be measured and evaluated to ensure these principles are adequately addressed. Furthermore, we also discuss the challenges and tradeoffs that could emerge by upholding these requirements in the healthcare domain. Since cardiovascular diseases represent one of the most active and high-impact areas in the development of data-driven medical models, we use this clinical domain to illustrate how the different TAI principles have been addressed in practice and to highlight the specific challenges encountered when applying them to medical AI systems in cardiovascular healthcare.
\end{abstract}



\end{frontmatter}



\section{Introduction}
\label{sec1}

The healthcare sector is under increasing pressure due to rising patient demands, demographic changes, resource constraints, and the complexity of medical decision-making. In this context, data-driven models powered by Artificial Intelligence (AI) and Machine Learning (ML) present a promising solution, offering the ability to extract valuable insights from the vast and continuously growing volume of medical data, ultimately enhancing efficiency, diagnosis, and treatment strategies \cite{alsalem_evaluation_2024}. Frequent applications of AI range from aiding in diagnosis and prognosis by predicting the progression of a disease and clinical events, providing input to clinical decision-making to establish new prevention and treatment strategies, intervention planning, or hospital resource planning \cite{gondocs_ai_2024}. However, the use of AI in the health sector remains in early stages, and most AI solutions do not extend beyond research contexts, with those implemented in actual clinical settings predominantly limited to biomedical research and administrative 'back office' applications \cite{hashiguchi_fulfilling_2022}. AI clinical data-driven models encounter several obstacles that hinder their practical implementation, such as ineffective deployment, complexities in data pipelines, and interoperability issues with health data \cite{ahmad_responsible_2023}. Furthermore, AI solutions can strain the physician-patient relationship by contributing to misinformation, potentially affecting clinical decision-making and communication \cite{szabo_clinicians_2022}. Additionally, AI-driven systems may introduce challenges that impact the patient experience, such as loss of privacy, heightened surveillance, increased inequality and discrimination, and the erosion of human autonomy and medical expertise. These challenges lead to ethical dilemmas and unintended consequences, which, in turn, erode user trust in AI solutions \cite{knopp_ai-enabled_2023}. Therefore, AI developers should learn to recognize and manage AI risks effectively \cite{markus_role_2021}. 

Policymakers and academics have been occupied with tackling the issue of trust and ethics in AI-based health applications, fostering the consideration of  trust as a design principle rather than an option \cite{albahri_systematic_2023, alsalem_evaluation_2024}, however, a fragile implementation of trustworthiness in AI solutions would exacerbate decision-making problems for patients and clinicians and weaken the accountability for errors, which would imply a critical barrier for their deployment in the clinical routine\cite{albahri_systematic_2023}. Embedding ethics and trust into the development of AI in clinical practice has proven difficult \cite{maris_ethical_2024}, due to the highly abstract nature of most guidelines that makes it unclear how to implement and regulate Trustworthy AI (TAI) systems in practice in the health domain. For instance, bringing the contextual understanding and clinicians' experience into AI models through a human-in-the-loop still remains a challenge \cite{onari_trustworthy_2023}, or, currently available explainability methods are mostly limited to static explanations, while recent works show that medical experts strongly prefer interactive explanations \cite{fehr_piloting_2022}. Additionally, a key point of this implementation is the procedure for evaluating completely different aspects of TAI through metrics that can provide objective and quantitative measurements of the medical AI system in certain critical scenarios. Nevertheless, trustworthiness is not solely determined by objective features; human judgment plays a significant role in factors such as usability. Therefore, the evaluation of trustworthiness should incorporate strict methodological procedures to handle subjectivity \cite{mattioli_towards_2023}.

Given the health sector's need for structured guidelines to align medical AI systems with the trustworthiness challenges arising from stakeholder interactions across various health processes \cite{zicari_co-design_2021}, it is essential to establish frameworks that integrate trustworthiness into the whole system’s life-cycle. Therefore, a trustworthy-by-design framework not only defines the key attributes an AI system must fulfill to meet healthcare stakeholders' expectations about trustworthiness, but also serves as a reference for AI developers to systematically implement trust-related aspects into the system.

This paper aligns with this vision by proposing a practical framework for ensuring trustworthy-by-design AI systems for the healthcare domain. Our framework can be adopted before the design of such AI systems, targeting two primary objectives: first, to reconcile the varying perspectives of health stakeholders on what constitutes TAI in the medical field; and second, to translate these diverse viewpoints into actionable and specific recommendations for developers of AI medical solutions, thereby enhancing stakeholder trust. Additionally, to demonstrate a practical application of the framework, we present a specific use case focused on data-driven models used for diagnosis and prognosis of cardiovascular diseases. As an overarching goal, we intend to shed some light on how to close the gap of non-clear implementation and regulation of TAI systems in the medical domain. Instead of focusing on a single trustworthy aspect, our contribution to the field is to present a comprehensive set of characteristics to be considered when designing AI medical systems that tackle all the trustworthy-related principles, namely: human agency and oversight, technical robustness, privacy and data governance, transparency, fairness, sustainability and accountability.  Additionally, this work seeks to address a series of research questions that will be explored through the proposed framework, including:
\begin{itemize}[leftmargin=*]
\item RQ1: What requirements must a medical AI system meet to comply with TAI principles? 
\item RQ2: Would healthcare stakeholders interact differently with these features, and if so, how?
\item RQ3: How can the fulfillment of these requirements be measured, and what key aspects should be assessed to ensure compliance?
\item RQ4: What strategies could be employed to manage the tradeoffs arising between different TAI principles?
\end{itemize}

The rest of the paper is structured as follows: first, Section 2 provides a comprehensive overview of TAI, including its definition, foundational concepts, core principles, and associated standards. Section 3 introduces the relevant entities within the healthcare sector, describing key healthcare processes and stakeholders. This section also presents a use case focused on the diagnosis and prognosis of cardiovascular diseases (CVD) to illustrate the framework's implementation. Section 4 lays the foundation for the proposed design framework, examining each TAI principle through a healthcare lens and outlining specific requirements to ensure trustworthiness from the design phase onward. For each principle, we also exemplify the implementation of our framework with respect to each TAI principle on the CVD use case. Section 5 explores potential tradeoffs between TAI principles that may arise during the framework's application, offering recommendations to manage them effectively. Section 6 identifies key challenges associated with the real-world implementation of the proposed framework in clinical settings. Finally, Section 7 presents the conclusions of our study, offering our responses to the aforementioned RQs based on insights drawn from preceding sections.

\section{Fundamentals of Trustworthy AI}\label{sec2}

Concerning medical AI solutions, the concept of ``trust'' inherent to TAI extends beyond its essential definition— assured reliance on the character, ability, strength, or truth of something—to include considerations from a human-technology interaction perspective necessitating a systematic, evidence-based approach that involves rigorous, standardized, and ethically grounded processes for design, validation, implementation, and monitoring \cite{chamola_review_2023}. Additionally,  trust in AI should not only be established between an AI system and the user, but also preserve the clinician-patient relationship characterized by trust, knowledge, regard, loyalty, and empathy \cite{drabiak_ai_2023}.

Thus, TAI can be defined as AI that empowers systems, devices and decision-making processes with reliable and secure mechanisms, ensuring they perform tasks safely, ethically, and consistently with the expectations of its audience. It involves enhancing privacy and security aspects, ensuring that AI systems are robust, transparent, and accountable \cite{cheng_socially_2021}.
Other related-concepts to TAI are currently used in the literature, but all of them tackle the aspects of ethical adherence, lawfulness, and technically robustness. For instance, \textit{Robust AI} regards AI systems with the ability to ``cope with errors during execution and cope with erroneous input'', \textit{Ethical AI} systems do what is right, fair, and just. Fair AI systems are absent from any prejudice or favoritism toward an individual or a group based on their inherent or acquired characteristics. \textit{Safe AI} envisages ways that do not harm humanity, while \textit{Dependable AI} focuses on reliability, verifiability, explainability and security. Finally, \textit{Human-centered AI} systems are continuously improving because of human input, while providing an effective experience between human and robots \cite{cheng_socially_2021}.

\subsection{Trustworthy AI principles}
\label{subsec22}

TAI is a dynamic, multidisciplinary field with no universally accepted definition or comprehensive guidelines outlining its implementation. This paper stems from the TAI principles described in the ethical guidelines for TAI delivered by High-Level Expert Group (HLEG) on AI of the European Commission \cite{doi/10.2759/346720}. This group proposes four ethical areas to achieve TAI such as human autonomy, prevention of harm, fairness, and applicability, which could be implemented through the following principles: 
\begin{enumerate}[leftmargin=*]
    \item \textbf{Human agency and oversight} which is crucial within the health sector as data-driven models seek to enhance clinicians' ability to make accurate, personalized, and timely decisions, without undermining their autonomy to make informed choices. It is essential to maintain stakeholder involvement in health decision-making through approaches that emphasize human oversight such as  Human-In-The-Loop (HITL) allowing users to intervene in the decision-making cycle, Human-On-The-Loop (HOTL) where the user participates in system design, and Human-In-Command (HIC)  to oversee overall AI activity and control its use \cite{el-sappagh_trustworthy_2023}.

    \item \textbf{Technical robustness and safety} emerge as critical in health domains since they address the data-driven model's sensitivity to input changes or adversarial attacks, such as data poisoning, model leakage or hardware/software attacks. Consequently, fallback plans and general safety measures should be taken to minimize associated risks. Additionally,  models must achieve high training and especially testing accuracies when operating in the health domain due to severe consequences of false positives and false negatives outcomes. Other aspects, such as reliability and reproducibility, must also be considered to ensure  the AI systems functions correctly across a variety of inputs and settings. 

    \item \textbf{Privacy and data governance} aim to protect sensitive patient information accessed by AI systems operating in healthcare environments. Additionally, the models' parameters and hyperparameters must be secured,  and any leaks during the data preprocessing phase must be prevented. The quality of the data impact directly on the model's performance and its integrity must be ensured before training or testing the model. Data access control is also crucial, with necessary protocols to determine who can access specific data and how.
    
    \item \textbf{Transparency}\ is crucial for achieving TAI models. Providing healthcare stakeholders with explanations for AI decisions is aligned with the \emph{right to explanation} under the European GDPR legislation \cite{schneeberger2020european}. Additionally, patients should be informed when they are interacting with AI models and have the option to choose between AI-generated and medical expert decisions. Furthermore, system decisions and the data used should be thoroughly documented to facilitate error tracking.

    \item \textbf{{Diversity, non-discrimination, and fairness}} are essential to ensure that all health stakeholders, whether directly or indirectly affected by the AI model, have equal access to its outputs. Any discrimination based on sensitive features not related to the medical problem should be anticipated and eliminated \cite{el-sappagh_trustworthy_2023}. In addition, considering the specific characteristics of the population targeted by the model, a one-size-fits-all approach should not be adopted, but rather universal design methods to accommodate all potential users should be utilized. 

    \item\textbf{Societal and environmental wellbeing} cover subprinciples such as sustainability and environmental friendliness, social impact, and democracy.  These subprinciples play a crucial role in ensuring that AI solutions in the health sector are not only effective but also accessible and beneficial to society as a whole while minimizing their negative impacts the environment. 

\item\textbf{Accountability} addresses three main components: auditability, enabling continuous assessment and audit of model functionality and the integrity of medical data by internal or external auditors; risk management to identify, assess, document, and mitigate potential adverse effects on patient outcomes; and redress mechanisms to swiftly address and rectify any erroneous diagnoses or treatment recommendations.
\end{enumerate}

\begin{figure}
    \centering
    \includegraphics[ width=1\linewidth]{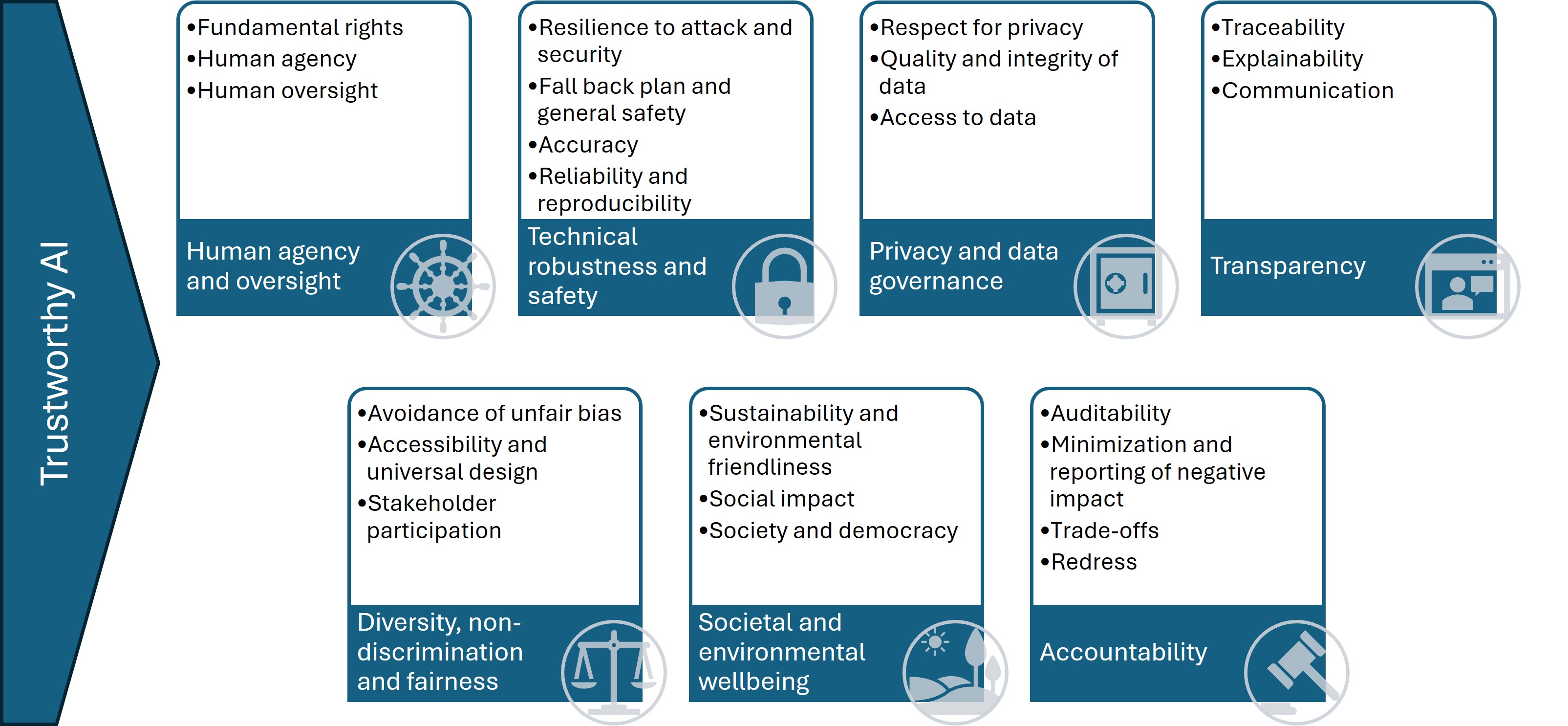}
    \caption{TAI principles and subprinciples according to the HLEG EU guidelines \cite{doi/10.2759/346720}.}
    \label{fig:TAI principles and subprinciples}
\end{figure}

Figure \ref{fig:TAI principles and subprinciples} represents the subprinciples included in the different principles described before.

\subsection{Standards associated with TAI in health}
\label{subsec23}

The implementation of these TAI principles will have a significant impact on how AI developers design, develop, and deploy data-driven models, which, particularly in the healthcare field, must adhere to certain standards and best practices. The first best practice to commit to could be the Hippocratic oath, which articulates principles still relevant today: 'I will respect all persons, without any discrimination,' 'I will inform patients,' etc. This alignment is not merely symbolic: AI systems deployed in medical settings increasingly act as decision-support agents with a direct influence on diagnoses, treatments, and patient outcomes. As such, they must be held to the same ethical standard as human practitioners. Just as physicians are expected to act in the best interest of their patients, AI systems must be developed and deployed with a commitment to non-maleficence, fairness, and transparency. Failing to embed these values from the outset can result in biased recommendations, lack of patient consent, or opaque decision-making, ultimately undermining trust and exacerbating health disparities. Upholding the spirit of the Hippocratic oath within AI system design should be the primal goal, to ensure that ethical responsibility must be preserved, regardless of whether decisions are made by a human or a machine.

In Europe, the EU Commission has established ethical principles for digital health, emphasizing humanistic values, individual data management, inclusivity, and eco-responsibility \cite{seroussi_implementation_2024}. These principles support the European Health Data Space, a key regulatory framework addressing both primary (healthcare provision) and secondary (research and policymaking) uses of health data \cite{hendolin2022towards}. Given AI’s reliance on personal data, the General Data Protection Regulation (GDPR) serves as a comprehensive legal framework ensuring explainability and accountability in automated decisions. A core GDPR objective is to protect individual rights in profiling and AI-driven decisions. While GDPR appears to mandate a uniform explanation model, in practice, explanation types—ex-ante, ex-post, expert or subject-oriented—vary based on context and the user’s ability to interpret the outcomes \cite{hamon_bridging_2022}.

Complementing GDPR, the AI Act ensures that medical AI systems comply with ethical and legal standards, addressing concerns related to bias, privacy risks, and patient safety \cite{aboy2024navigating}. Proposed by the European Commission, the AI Act establishes a regulatory framework prioritizing safety, transparency, fairness, and accountability in AI applications. It classifies AI systems by risk, imposing strict requirements—such as data governance, human oversight, and transparency—on high-risk sectors like healthcare. By mandating conformity assessments and continuous monitoring, the AI Act fosters trust in AI-driven medical solutions, reinforcing a trustworthy-by-design approach that enhances clinical decision-making while safeguarding patient rights.

Several international frameworks and standards promote TAI across various sectors, including healthcare. The ISO/IEC 42001:2023 AI Management System Standard outlines requirements for ethical AI development, risk management, and data governance \cite{benraouane2024ai}. The NIST AI Risk Management Framework (AI RMF) provides guidelines to mitigate AI-related risks by enhancing fairness, transparency, security, and accountability \cite{ai2023artificial}. The World Health Organization (WHO) Guidance on AI Ethics and Governance ensures AI applications in healthcare align with ethical, transparent, and accountable practices \cite{guidance2021ethics}. Collectively, these frameworks contribute to the global effort in establishing robust AI governance for responsible development and deployment.


\section{An overview of the AI ecosystem in the medical field: processes, data, and stakeholders}
\label{sec3}

To design TAI systems in the medical domain, it is essential to first understand the broader ecosystem in which these technologies operate. Before introducing the proposed design framework, this section offers an overview of the key components that define the medical AI landscape, namely, the processes involved (Subsection \ref{subsec31}), the diverse types and sources of medical data (Subsection \ref{subsec32}), and the various stakeholders who interact with AI systems at different stages of their lifecycle (Subsection \ref{subsec33}), together with the ways in which such stakeholders engage with AI technologies across the healthcare value chain (Subsection \ref{subsec34}). Mapping this ecosystem is crucial for grounding the discussion in the realities of medical practice and ensuring that trustworthiness is addressed in context. Subsection \ref{subsec34} also introduces a use case on TAI for cardiovascular diseases that will serve as an example of the application of the design framework described subsequently in the rest of the manuscript. 

\subsection{Medical processes}
\label{subsec31}

Within the health domain, various processes represent specific stages or activities in the care delivery system, organized around the needs of patients and the requirements of clinicians and healthcare providers:
\begin{itemize}[leftmargin=*]
\item \textit{Screening} involves medical tests or procedures performed on members of a defined asymptomatic population to assess the likelihood of having a particular disease, with early detection being a major objective to increase treatment success \cite{maxim_screening_2014}.  

\item \textit{Diagnosis} is the process used by clinicians to identify the nature and cause of a patient’s symptoms through the evaluation of medical history and physical examination, which is crucial for effective treatment planning and management \cite{maxim_screening_2014}.

\item \textit{Prognosis} predicts the likely course and outcome of a disease, aiding in future treatment planning and giving patients realistic expectations about their recovery or condition management. \textit{Treatment} refers to the management and care of a patient for combating a disease, condition, or injury, involving a wide range of activities such as administering medications, performing surgeries, recommending therapeutic exercises, and implementing lifestyle changes \cite{loscalzo2022harrison}. 

\item Additionally, \textup{disease monitoring} or \textit{follow-up} is a critical aspect of medical treatment that ensures the healthcare team continues to monitor the patient’s progress even after the active treatment phase ends \cite{kimman_follow-up_2022}.  
\end{itemize}

\subsection{Data types in medicine}
\label{subsec32}

Data is the foundation of medical AI, where its sensitivity, confidentiality, and interoperability pose unique challenges. Medical AI systems process various data types, for example, medical notes (unstructured clinician narratives documenting patient interactions and care plans) \cite{wang_accelerating_2023}, clinical tabular data (structured EHR information, including demographics, medications, and lab results) \cite{murdoch2013inevitable}, medical images (X-rays, MRIs, CT scans, and ultrasounds critical for diagnosis) \cite{litjens2017survey}, biomedical signals (time-series data like ECGs and EEGs for real-time monitoring) \cite{bajaj2021biomedical}, and genomics (DNA analysis for precision medicine and personalized treatments) \cite{snyder2016genomics}. Effective AI analysis depends on tailoring methods to each data type, ensuring accuracy and reliability, particularly for high-dimensional or unstructured data, which may contain irrelevant or redundant features, leading to overfitting and poor generalization if not properly handled.

Beyond predictive performance, high-quality medical data is crucial for ensuring accountability and traceability in AI-assisted decisions. In safety-critical environments like healthcare, reliable data enables proper postmortem analysis when errors occur, supporting transparency and continuous improvement. Moreover, robust data foundations allow for the identification of sources of failure and support redress mechanisms, which are key to maintaining patient trust and meeting legal and ethical standards for responsible AI deployment. To this end, robust data governance mechanisms are crucial for supporting the responsible use of AI systems in the medical domain. These include policies and practices for data quality assurance, access control, auditability, provenance tracking, and compliance with privacy regulations. 

\subsection{Healthcare stakeholders}
\label{subsec33}

In the design phase of AI systems for medicine and healthcare, involving key stakeholders is essential to ensure that TAI principles align with their diverse needs, thereby preventing significant misalignment in problem focus, data utilization, feature analysis, and metric prioritization \cite{hashiguchi_fulfilling_2022}. Thus, achieving TAI requires a holistic and systemic approach that encompasses the trustworthiness of all actors and processes within the AI system's socio-technical context. From a domain-agnostic perspective, Barclay et al. 2021 \cite{barclay_identifying_2021} propose four general roles -- data scientists, ML engineers, system integrators, and domain practitioners --who interact through various machine learning and system integration processes involving datasets, ML models, and AI systems.  The medical domain necessitates a more tailored approach that recognizes the unique needs and nuances of its domain practitioners and processes to address the principles of TAI effectively \cite{saw_current_2022}. Besides patients, among the stakeholders involved in the lifecycle of AI systems for healthcare are those who interact directly with patients, such as clinicians and healthcare workers, as well as those working ``behind the scenes'' like regulators, policymakers, and healthcare providers (hospitals and care centers), each recognizing AI demands from different perspectives. Also integral to this group are AI developers, including data scientists and system developers, who are responsible for creating the necessary algorithms and infrastructure.\cite{concannon_practical_2019}.

\subsection{AI interaction with healthcare stakeholders}
\label{subsec34}
\begin{figure}
    \centering
    \includegraphics[ width=1\linewidth]{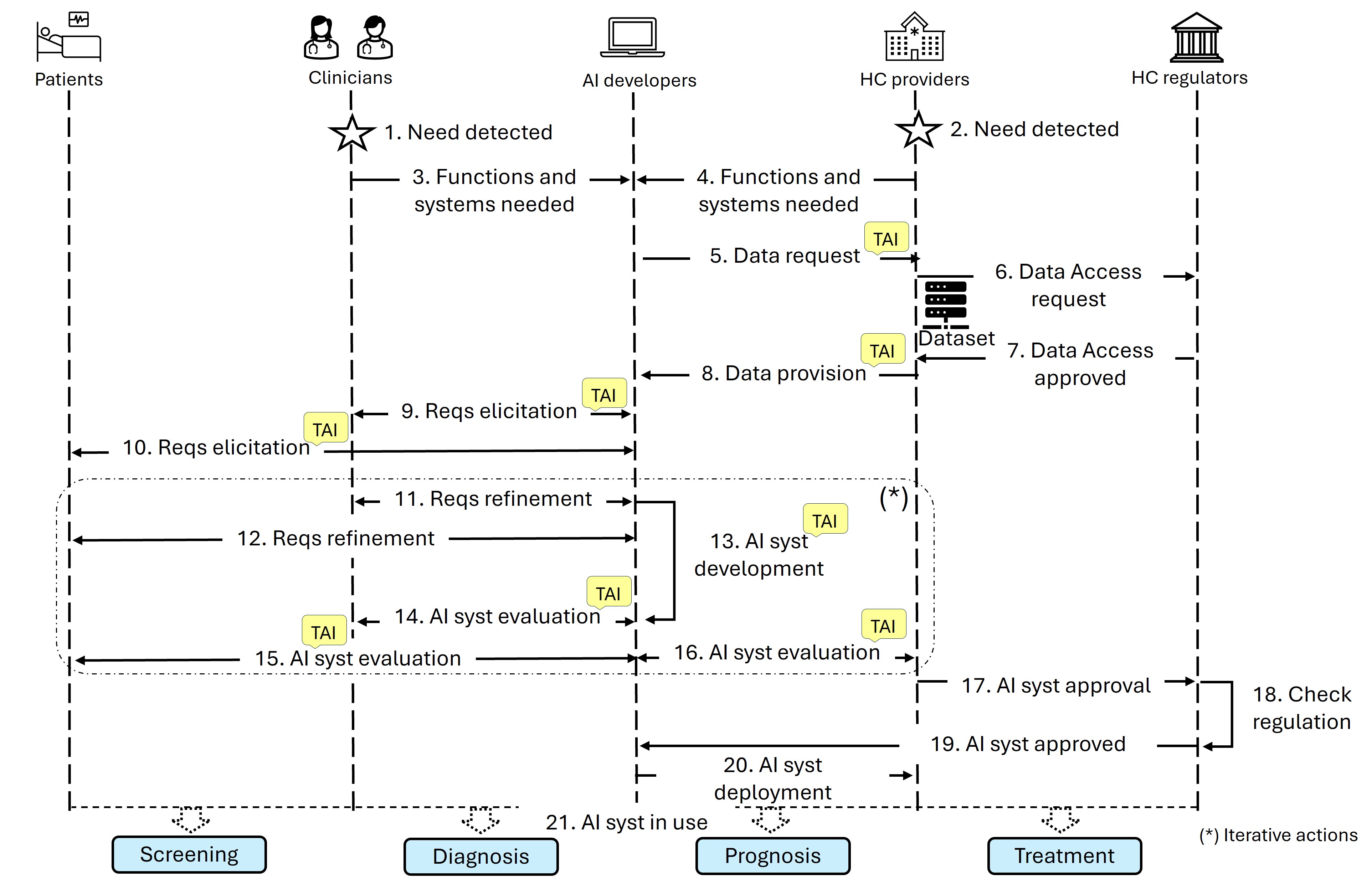}
    \caption{Healthcare stakeholders interaction during design and development of AI medical system. HC: Healthcare. (*): iterative actions}
    \label{fig:Design&Dev}
\end{figure}

Similar to other trust networks within the healthcare system, trust in AI must be established not only between the AI system and its users, but also between stakeholders involved in the process. Therefore, it requires a clear understanding of the interaction of AI system with various stakeholders to effectively address all parties involved in achieving TAI. This section illustrates the interactions an AI tool would have across various health processes, involving different stakeholders and governed by the framework proposed in this paper. Before deploying the AI system in a medical process, the design and development phases are crucial for integrating TAI aspects, as depicted in Figure \ref{fig:Design&Dev}.

Initially, after the needs are reported by the AI system’s potential users (steps 1-4 in Figure \ref{fig:Design&Dev}), the collection and provision of data for model development and training, which has to be approved by regulators, must comply with privacy and fairness principles (steps 5-8). Furthermore, during the elicitation of user and system requirements (steps 9-10) and their subsequent refinement (steps 11-12), AI developers must ensure that all TAI principles are addressed to mitigate any ethical concerns. During both development and evaluation phases (steps 13-16), the TAI principles defined during the design phase should be cross-checked in collaboration with stakeholders affected by the system's decisions, i.e., clinicians, Healthcare (HC) providers and patients´organizations, which act as proxies representing the patients´ perspective. Before the AI-based tool is deployed,  all regulatory requirements are met (steps 17-20) to ensure the AI system becomes ready for use in any of the aforementioned health processes. 

\begin{figure}
    \centering
    \includegraphics[ width=1\linewidth]{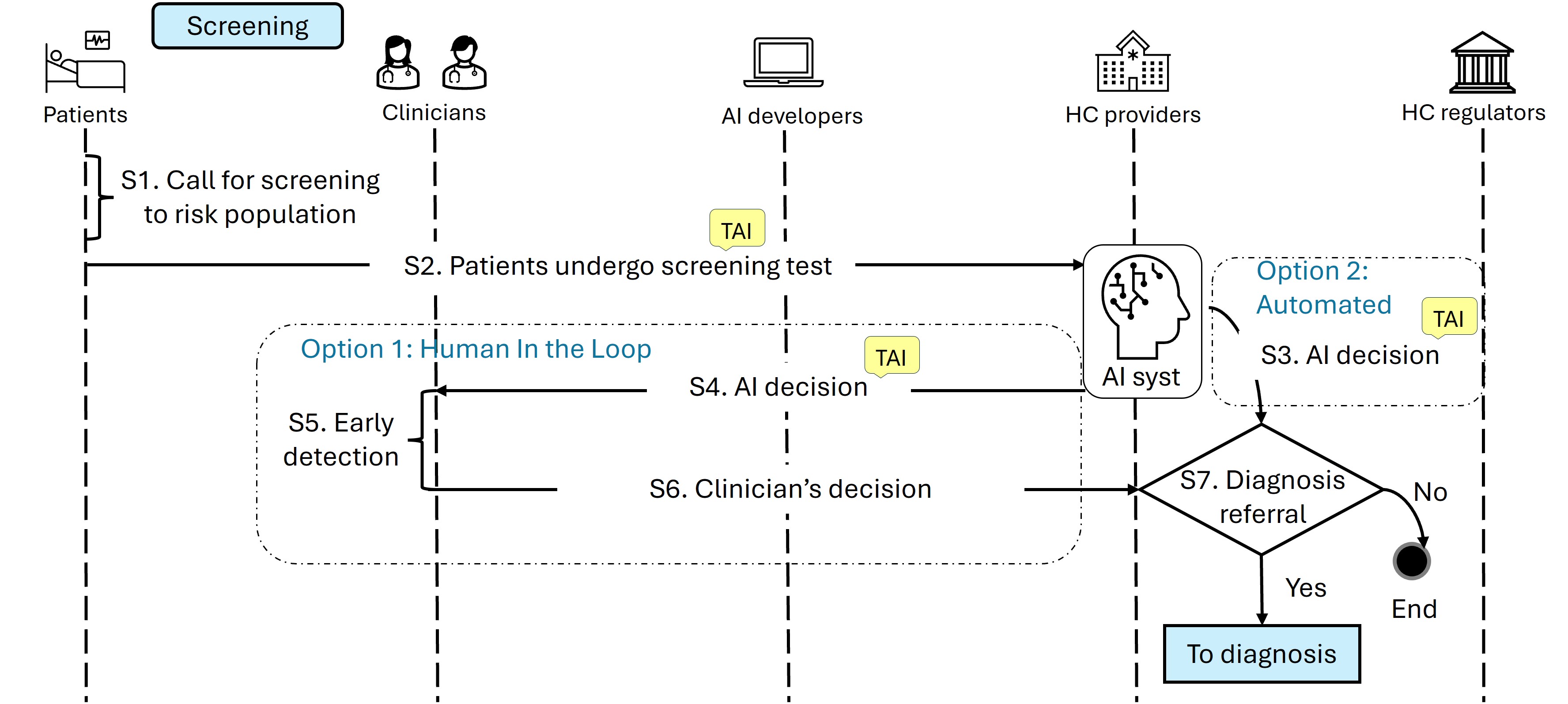}
    \caption{Healthcare stakeholders interaction during screening process.}
    \label{fig:Screening}
\end{figure}

During the screening phase, shown in Figure \ref{fig:Screening}, any interaction with the AI system by either patients or clinicians must adhere to TAI principles (steps S2 and S4). The decision made by the AI-based tool, which should inform about the probability of an existing disease along with confidence intervals, can be operationalized in two ways: either through a Human-in-the-Loop/Human-in-Command approach, where the doctor's decision is informed and supported by the AI tool (steps S4-S6), or through an automated decision process (step S3) that directly leads to a possible diagnosis referral (step S7). In the latter case, special attention must be given to human agency and accountability. It is important to note that, particularly during the screening phase, individuals are not yet patients in the clinical sense, but rather citizens undergoing preventive assessments who may potentially be identified as patients based on the test outcomes."

\begin{figure}
    \centering
    \includegraphics[ width=1\linewidth]{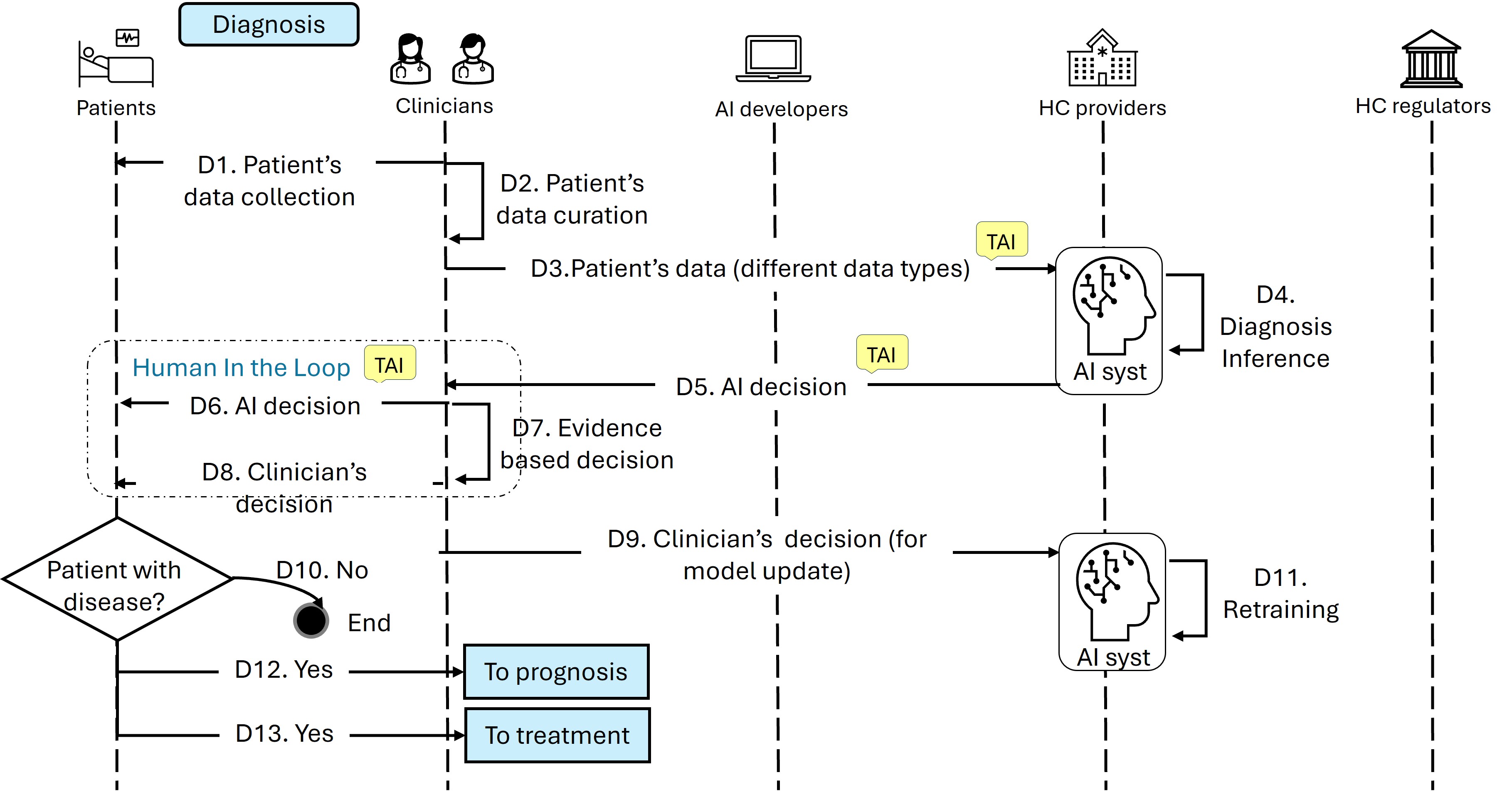}
    \caption{Healthcare stakeholders interaction during diagnosis process.}
    \label{fig:Diagnosis}
\end{figure}

The diagnosis process, depicted in Figure \ref{fig:Diagnosis},  can be initiated by a clinician or follow up a screening procedure (Step S7). After the collection of patient data, clinicians carefully curate the information before forwarding it to the AI-based system, which is conducted in alignment with TAI principles (Steps D1-D4) . In this stage, a human-in-the-loop approach is essential, as the doctor cross-references the AI's decision with their own expertise and clinical evidence to formulate the final diagnosis, then leading to the subsequent prognosis and/or treatment steps (Steps D5, D6, D7, D8, D12, and D13). The doctor's final decision is also utilized to update the model parameters for future decisions (Steps D9 and D11).

\begin{figure}
    \centering
    \includegraphics[ width=1\linewidth]{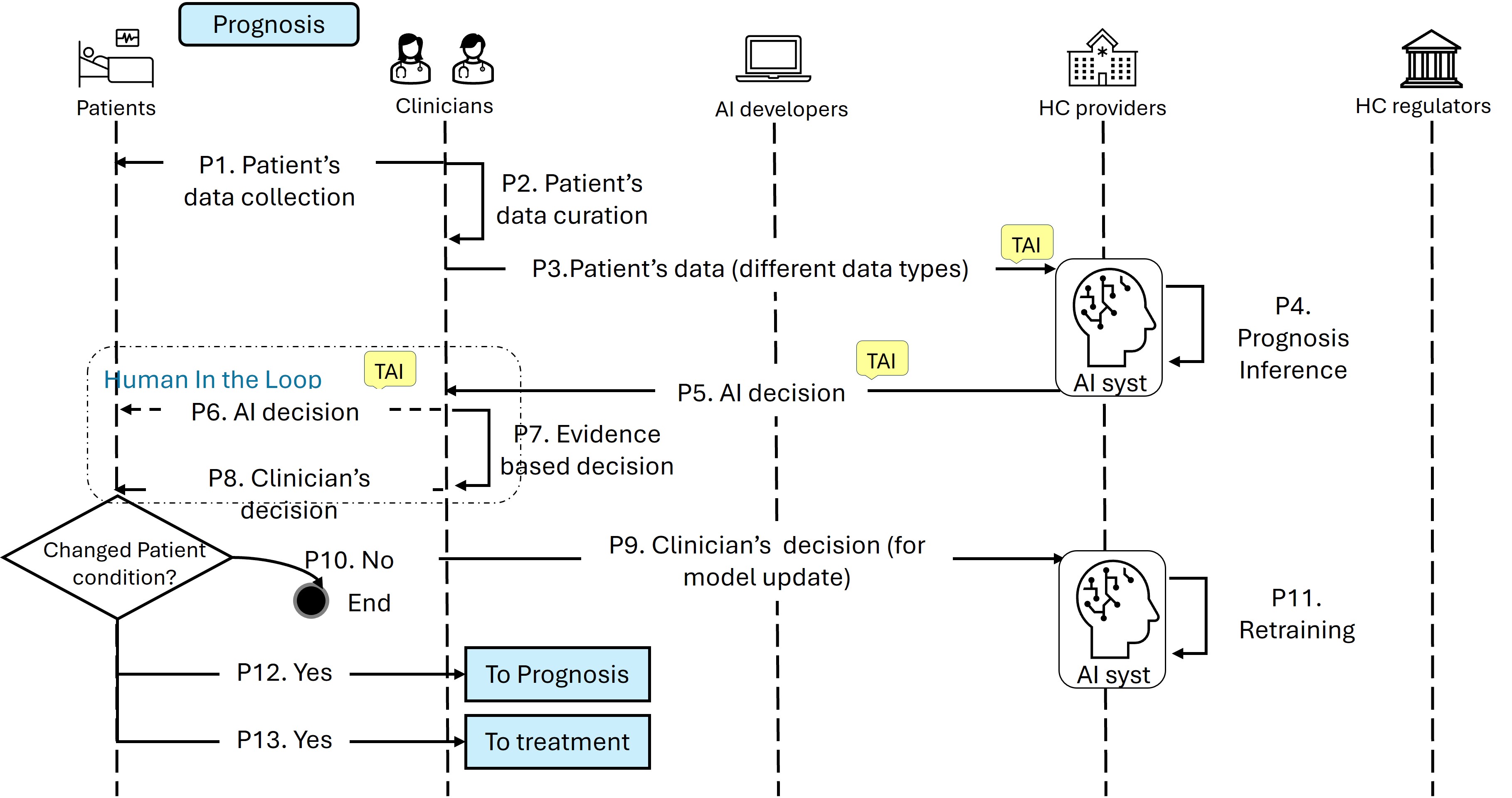}
    \caption{Healthcare stakeholders interaction during prognosis process.}
    \label{fig:Prognosis}
\end{figure}

The prognosis process, shown in Figure \ref{fig:Prognosis}, has similarities to the diagnosis process concerning user interactions. However, this process may recur periodically if the patient's condition changes and an adjustment of the prognosis is required (P12) or a new treatment has to be conducted (P13).

\begin{figure}
    \centering
    \includegraphics[ width=1\linewidth]{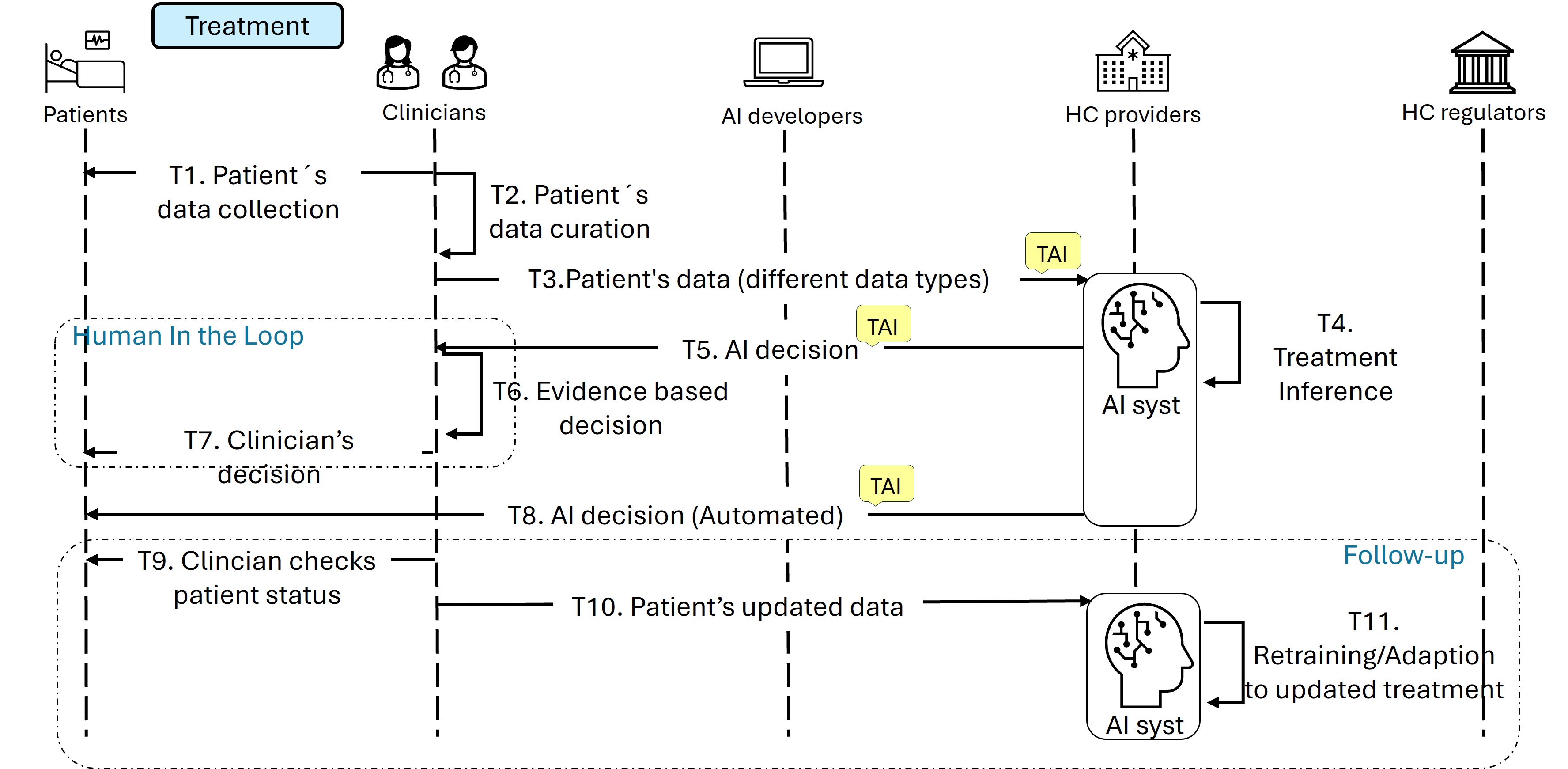}
    \caption{Healthcare stakeholders interaction during treatment process.}
    \label{fig:Treatment}
\end{figure}

Likewise, the treatment process, Figure \ref{fig:Treatment},  can involve iterative cycles where treatment actions are revised and updated in a further follow-up or monitoring of the patient status. The interactions with the AI system by users, whether patients or HC professionals, are subject to TAI principles (Steps T3 and T4). Additionally, the treatment process may include an automated decision directly provided to the patient (Step T8). During the follow-up, the model may be updated or retrained by the HC professional´s request to refine future decisions based on the patient's progress during treatment (Steps T9-T11).

\begin{figure}
    \centering
    \includegraphics[width=\textwidth]{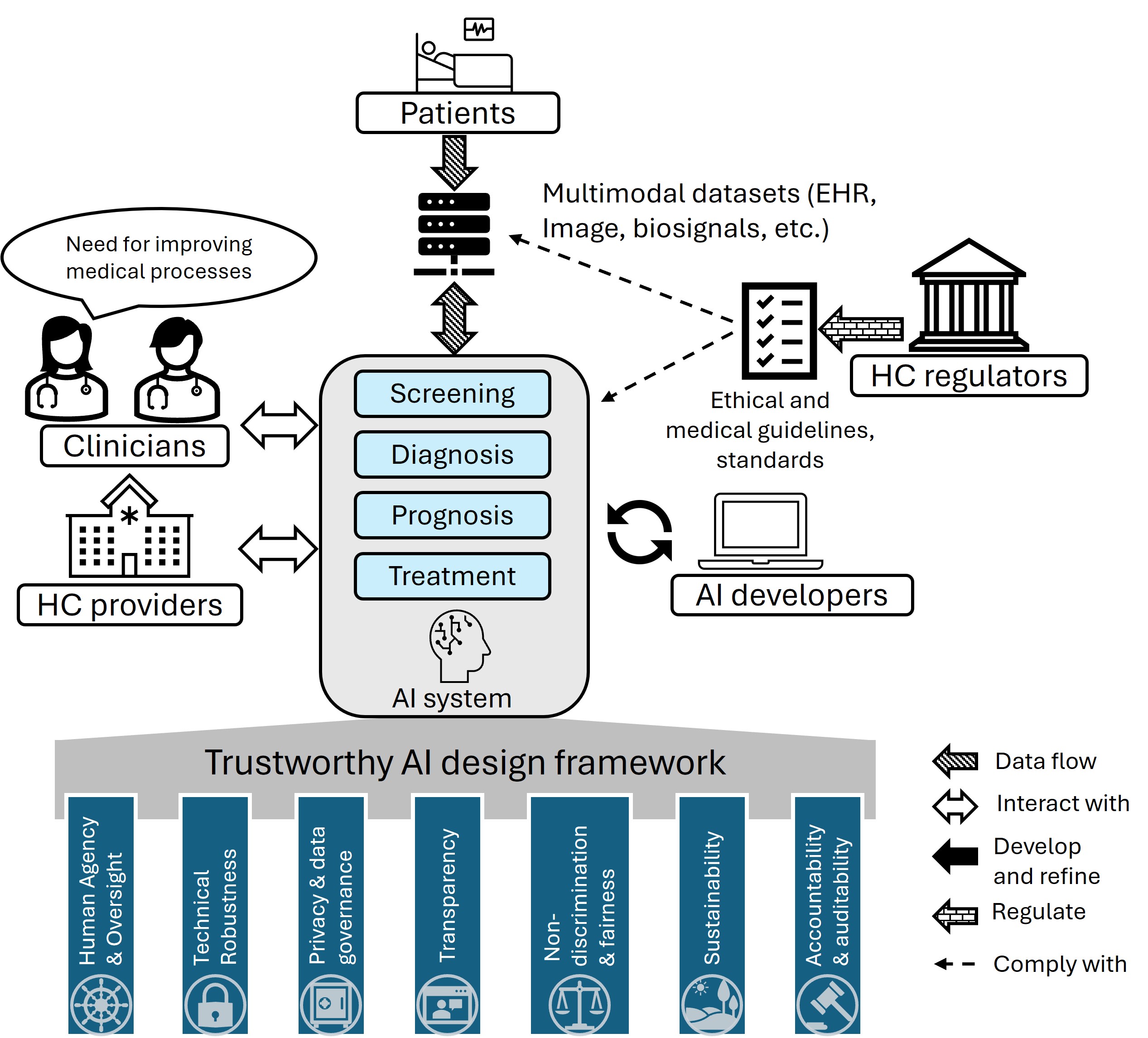} 
    \caption{Overview of stakeholders interactions within the TAI design framework.}
    \label{fig:TAI_framework_structure}
\end{figure}

Figure \ref{fig:TAI_framework_structure} illustrates how health stakeholders interact with the AI system, emphasizing the inherently collaborative nature of healthcare processes involving patients, clinicians, healthcare providers, and policymakers, as well as various data modalities such as electronic health records (EHR), physiological signals, and medical images. The figure also demonstrates how the AI system integrates the TAI design framework, highlighting the alignment of system functionalities with principles such as transparency, accountability, and human oversight. Among clinicians, it is essential to consider a broad spectrum that ranges from general practitioners in primary care to specialists in different medical fields, as well as other supporting medical staff, including nurses. Healthcare providers encompass not only hospitals as specialized care centers but also primary care facilities and occupational health organizations. Meanwhile, policy bodies play a crucial role in ensuring compliance with medical guidelines, including professional societies and hospital ethics committees, as well as regulatory entities overseeing data privacy and AI governance, such as GDPR and AI Act enforcement agencies, and standardization bodies like ISO and CEN.


\paragraph{Use case: Trustworthy AI in cardiovascular diseases} As one of the leading causes of death globally, Cardiovascular Diseases (CVD) account for approximately 17.9 million deaths each year, representing 32\% of all global mortality \cite{amini2021trend}. Given the proven utility of AI in supporting diagnosis and prognosis within the CVD field \cite{sun2023artificial}, ensuring that such systems adhere to TAI principles is essential to promote clinician and patient acceptance and safe deployment in real-world clinical settings. However, the existing literature on AI-based CVD solutions often lacks a comprehensive consideration of trustworthiness throughout the AI lifecycle \cite{moreno2024ecg}. While this paper focuses broadly on TAI across different medical processes, we use the CVD field illustratively to reflect the relevance and practical challenges of operationalizing TAI. In the subsequent development of the requirements for each TAI principle (Section \ref{sec5}), we point to recent works that have attempted to operationalize these principles in medical AI systems for CVD, highlighting their strengths, limitations, and the ongoing need for structured, principle-aligned development.


\section{Design framework for medical AI systems}
\label{sec4}

In this section, we take a deeper look into the TAI (TAI) principles and their specific impact on healthcare. For each principle, we begin by presenting relevant definitions and concepts to facilitate a clear understanding of the requirements that follow. Next, we enumerate and describe, in tabular form, the features, recommendations or requirements that an AI solution for healthcare should achieve to fulfill each specific TAI principle and its sub-principles, considering the perspectives of various health stakeholders. Since some features may be relevant across multiple stakeholders, we provide a matrix to map the involvement of each stakeholder with the defined requirements. The proposed categorization of stakeholder relationships to the requirements that makes the AI system trustworthy are \emph{Responsible} (R), \emph{Affected} (A), \emph{Uses} (U), \emph{Monitors} (M), \emph{Contributes} (C), and \emph{Informed} (I). Stakeholders who are \textit{Responsible} bear accountability for ensuring that a requirement is implemented correctly and maintained, such as healthcare providers overseeing data privacy compliance. Those who are \textit{Affected} experience direct or indirect consequences of how the requirement is addressed, like patients impacted by biased algorithms. Stakeholders who \textit{Use} a requirement actively engage with its functionality, such as clinicians leveraging explainability features for decision-making. Those who \textit{Monitor} are tasked with verifying that requirements are continuously met, ensuring adherence to standards and ethical guidelines, as seen with healthcare regulators. Stakeholders who \textit{Contribute} provide essential input or feedback during the requirement’s design or refinement, such as clinicians participating in user-testing phases. Lastly, those categorized as \textit{Informed} need to stay updated about the requirement’s implementation and outcomes, like patients being notified of privacy practices. Additionally, we assign a qualitative priority score based on the MoSCoW methodology \cite{miranda2022moscow} that suggests whether a requirement must (M), should (S) or could (C) be imposed. We also present an overview of the aspects within each TAI principle that should be evaluated. Finally, we orient the requirements proposed to the application in the use case for diagnosis and prognosis of CVDs.

\subsection{Fundamental rights, human agency and oversight}
\label{subsec41}

When addressing the principle of ``Fundamental rights, human agency, and oversight" in medical AI systems, developers should adopt the Human-Centered AI (HCAI) paradigm during the design phase. HCAI empowers users by providing fair, transparent access and control over both data and the algorithms underlying AI systems \cite{nagitta_human-centered_2022}. 

Fundamental rights are central to any AI system designed to interact with or impact humans. In medical practice, AI should not harm or compromise human dignity or integrity, either mentally or physically. According to Dlugatch et al. and Maris et al., all patient care decisions must be made by humans, with AI only providing support, ensuring that humans remain responsible for the final decision at all times \cite{dlugatch_trustworthy_2023, maris_ethical_2024}.  Patient consent is widely recognized as a fundamental right in the ethical use of personal health data, grounded in the principles of autonomy, dignity, and informational self-determination. This right is especially important in the context of AI systems that rely on large-scale data for training and validation. However, regulatory frameworks may permit the use of anonymized health data for secondary purposes—such as research, quality control, or model development—without explicit patient consent, as long as strict data protection measures are in place. For instance, in countries like Finland \cite{bruck2024european}, patients are not required to consent to the use of fully anonymized data for these purposes. Nevertheless, the right to transparency and control is preserved: patients can request to know how their data is being used and may demand its removal from specific research projects. This underscores the need for mechanisms that support patient agency and uphold fundamental rights, even when consent is not legally mandated.

In the context of healthcare, agency refers to the ability of clinicians to make autonomous, informed decisions. However, many clinicians are concerned about losing this autonomy due to AI's promising capabilities. To address this, AI should be used to enhance their agency, acting as a ``colleague or agent'' that supports decision making by highlighting overlooked features or potential flaws in judgment \cite{zou2025rise}. While this approach can strengthen expert decision-making, it also carries risks, such as overreliance on AI, which could lead to deskilling and a significant loss of medical expertise \cite{gondocs_ai_2024}.   

Clinicians agency should be closely linked with oversight, ensuring they remain part of the clinical process, commonly referred to as human-in-the-loop, to maintain an active role in patient care, including the ability to override AI decisions \cite{buruk_critical_2020}. Human oversight can be integrated throughout all phases of an AI system’s lifecycle: HOTL is suited for planning and design, HITL for data collection, modeling, and deployment, and HIC for monitoring overall system use \cite{kaur_trustworthy_2022}. These oversight approaches introduce an essential layer of ``causal'' intelligence to medical AI systems \cite{muller_explainability_2022}, as clinicians use their experience to interpret model correlations, infer underlying cause-effect relationships, and compensate for the lack of causal guarantees in most AI systems, thus reinforcing the need for human involvement and promoting agency and oversight. The features and requirements that we proposed to embrace this principle are listed in Table \ref{tab:Human agency}.

The procedure to measure the fulfillment of this principle would rely on a qualitative evaluation that addresses each of the areas of the principle. A risk assessment focused on fundamental rights (e.g., ensuring transparency in informed consent) is an effective approach to protecting patient dignity and integrity. Regarding human agency, we propose assessing various dimensions of agency as outlined by Dattathrani et al. \cite{dattathrani2023concept}, including passivity (being acted upon rather than acting), automaticity (responding to stimuli unconsciously or habitually), rationality (acting based on reasons, beliefs, and intentions), endorsement (involvement in planning and executing actions), freedom-to-choose (awareness and freedom to select from multiple motivations), and consciousness (ability to store, report, and integrate mental states to guide action). Evaluating human oversight in medical AI systems requires focusing on critical clinical decision points. Oversight mechanisms must be clearly documented and stakeholders need defined roles to actively intervene or override AI decisions during critical care processes. To this end, scenario-based testing assesses whether clinicians can effectively intervene under real-world clinical conditions, which can be combined with satisfaction surveys to assess if they feel empowered to act when necessary. Other indicators such as intervention frequency and response speed can measure the effectiveness of oversight.

\begin{table}[h!]
\centering
\caption{Requirements for the first TAI principle: fundamental rights, human agency and oversight.}
\label{tab:Human agency}
\vspace{2mm}
\resizebox{\textwidth}{!}{%
\begin{tabular}{cm{3cm}m{6cm}ccccccc}
\toprule
\textbf{Id} & \textbf{Subprinciple} & \textbf{Requirement} & \makecell{\textbf{Priority}\\\textbf{scale}} & \makecell{\textbf{Medical}\\\textbf{process}} & \textbf{Patient} & \textbf{Clinician} & \makecell{\textbf{HC}\\\textbf{Provider}} & \makecell{\textbf{HC}\\ \textbf{Regulator}}  & \makecell{\textbf{AI}\\\textbf{developer}}\\ \midrule
R1.1 & Fundamental rights & \raggedright Patients are provided with mechanisms to request information and exercise control over the use of their data within the AI system (e.g., for secondary use, model training, or decision inference)& Must
& All
& A, U, R& I&I, M& I&C
\\ \midrule
R1.2 & Fundamental rights, Human agency& \raggedright The patient is provided with a mechanism to withdraw consent, allowing them to stop the AI system from using their data in any future training& Should& All& A, U, R& I& I, M&  I&C
\\ \midrule

R1.3 & Fundamental rights, Human agency& \raggedright The clinician maintains autonomy and is not subject to any form of manipulation when using the AI system& Must& All&  I& A, U, R& I, M&  I&C
\\ \midrule

R1.4 & Fundamental rights, Human agency& \raggedright The clinician is provided with detailed information about all actions that can be performed using the AI system& Should& All& I& A, U, R& I, M&  I&C
\\ \midrule

R1.5 & Human agency & \raggedright The AI system offers general information about its features to allow stakeholders to understand its functioning &Should &All &U& U&U, M&U&R
\\ \midrule

R1.6 & Human agency & \raggedright The AI system offers a mechanism to support independent decision-making by clinicians, acting as a recommender while ensuring the final decision is made by the clinician
 & Must &All & I& U, A& I, M&  I&R
\\ \midrule

R1.7 & Human agency
 & \raggedright The patient is provided with a mechanism to consent to whether the clinician's decision is supported by the AI system's output
 &Could &All & U, R&	A&	I, M&	I&	C
\\ \midrule

R1.8 &Human agency & \raggedright  Despite AI systems´ high performance,  the clinician retain final decision-making authority, minimizing risk of automation bias and overreliance on AI´s outputs
 &Must &All &A, I&	U, R& 	I, M&	I&	C
\\ \midrule

R1.9 & Human oversight & \raggedright The AI system allows stakeholders to interact with the model to inspect the decision-making steps and understand the basis for AI-generated recommendations
 &Should &All & A, U&	U&	U, M&	I&	R
\\ \midrule

R1.10 & Human oversight & \raggedright The AI system allows stakeholders to intervene decision support-steps (modify inputs, model´s weight and hyperparamenters) to explore new ad-hoc uses.
 &Could &All &A, I&	U&	U, M&	U&	R
\\ \midrule

R1.11 & Human oversight & \raggedright A Human-in-Command and/or Human-in-the-Loop mode is enabled by default,  allowing users to decide when to apply AI system outputs to a specific case
 &Must &All &A, I&	U&	U, M&	I&	R
\\ \bottomrule 
\multicolumn{10}{l}{$\ast$ R: Responsible, A: Affected, U: Use, M: Monitor, C: Contribute, I: Informed, HC: Healthcare} \\
\end{tabular}
}
\end{table}

\paragraph{Use Case: TAI in CVD} AI applications in cardiovascular care must augment rather than replace human clinicians. Studies emphasize that patients are more likely to trust AI tools when physicians remain actively involved and ultimately responsible for decision-making \cite{Mooghali2024}. For example, automated ECG interpretation systems function as assistive tools, with cardiologists retaining final judgment to prevent overreliance and preserve patient autonomy. However, putting this principle into practice poses challenges, including increased clinician workload and the need for training. A sudden technology failure or an algorithm’s blind spot can jeopardize care if humans are unprepared to intervene \cite{Niroda2025}. To address this, researchers recommend comprehensive clinician training, clear accountability protocols, and human-in-the-loop designs—such as decision support systems that flag high-risk cases for review. These approaches ensure that AI strengthens, rather than diminishes, human agency across cardiovascular screening, diagnosis, and treatment.

\subsection {Technical robustness and safety}
\label{subsec42}

Recently, AI development in healthcare has often neglected safety issues, as models were primarily intended for research rather than real patient use \cite{oprescu_towards_2022}. Today, safety must be prioritized from the design phase onward. Information security management now demands a comprehensive risk analysis that integrates technology, human factors, and the scope of healthcare services. Robustness addresses safety by ensuring the model remains accurate with unseen data, a critical factor in healthcare due to potential patient safety impacts \cite{onari_trustworthy_2023}. In healthcare, robustness extends beyond cross-validation and feature engineering; it involves avoiding overfitting and focusing on medically significant features to ensure consistent diagnoses, even with small input variations and missing data\cite{karim_adversary-aware_2022}. 

The handling of missing data is essential to maintain technical robustness and safety in medical AI systems, as incomplete clinical records (resulting from inconsistent documentation, fragmented care pathways, or unavailable diagnostic tests) can introduce bias and compromise the reliability of predictions. Several strategies are available to address this issue. Traditional imputation methods, such as mean substitution, regression imputation, and multiple imputation (including multiple imputation by chained equations, MICE), estimate missing values based on observed data patterns \cite{afkanpour2024identify, arnaud2023predictive}. Alternatively, simpler approaches like case-wise deletion (removing patients with missing values) may be applied, although this risks reducing sample size and representativeness. More advanced solutions, including deep learning-based imputation models, are increasingly used to capture complex relationships in high-dimensional medical data \cite{liu2023handling}. To ensure methodological transparency, researchers are encouraged to clearly report the extent of missing data and the techniques used to manage it. In addition, the choice of imputation strategy should be guided by the underlying mechanism of missingness: whether data are completely missing at random (MCAR), at random (MAR), or not at random (MNAR) and by the specific clinical and analytic context of the study.  

Designing robust medical AI requires a proactive approach to risks to ensure reliable operation under dynamic conditions while mitigating unintended behaviors \cite{chamola_review_2023}. To prevent performance risks, identifying non-robust features is crucial, as they may lead to inaccurate or harmful predictions for patients \cite{guesmi_physical_2023}. A similar approach to improving classifier accuracy in differential diagnosis uses debias training, where clinicians blacklist irrelevant features to prevent faulty attributions \cite{lu_building_2023}. Accuracy also depends on expert consensus regarding the suitability of clinical data used in the model, especially in health applications affecting patients \cite{massella_regulatory_2022}. 
The most effective way to address issues of accuracy, robustness, and generalizability is through extensive external validation, though this can be slow due to lack of standardization. In healthcare, the OHDSI (Observational Health Data Sciences and Informatics) network's standards, like OMOP-CDM (Observational Medical Outcomes Partnership-Common Data Model), facilitate large-scale external validation of patient-level prediction models, ensuring reproducibility \cite{ahmadi2022omop}.

Robustness risks in medical AI arise from both internal system performance and external adversarial attacks, leading to flawed clinical recommendations. Adversarial attacks can be categorized into four types: adversarial falsification (e.g., data poisoning through false positives/negatives), adversarial knowledge (white-box and black-box attacks), adversarial specificity (targeted and non-targeted misclassifications), and attack location (e.g., poisoning during training or backdoor attacks) \cite{guesmi_physical_2023, karim_adversary-aware_2022}. Due to the complexity of medical inputs (e.g., images, biosignals) and the possibility of observing unusual symptomatic disease patterns, detecting out-of-distribution (OOD) data is challenging, impacting model robustness and clinical decision-making. To address these vulnerabilities in medical AI systems, both proactive and reactive measures are crucial throughout the lifecycle. Proactive strategies during development include adversarial retraining, and generating realistic adversarial examples --considering the operational domain in which the AI tool is to be used \cite{herrera2025responsible}-- to ensure the model's reliability in clinical decision-making. Reactive measures, applied post-deployment, involve anomaly detection, adversarial detection, and input reconstruction, all aimed at maintaining robustness to safeguard patient outcomes and ensure trustworthy medical recommendations \cite{goodfellow_explaining_2014}. Table \ref{tab:Robustness} outlines the key features and requirements identified to support the implementation of this principle.

To effectively ensure robustness, models need thorough evaluation. When facing adversarial attacks, robustness can be assessed at two levels: local robustness, which checks if a model's output remains stable under small perturbations, and global robustness, which evaluates consistency across an entire dataset under perturbations \cite{mattioli_towards_2023}. Li et al. use calibration plots and Brier scores to measure discrepancies between predicted and observed outcomes \cite{li_using_2023}. Göllner and Tropmann-Frick propose several robustness measures, such as zeroth-order optimization for tabular models, and FGSM, PGD, and DeepFool for computer vision, as well as TextBugger and PWWS for NLP models \cite{gollner_bridging_2023}. Toolkits like IBM Adversarial Robustness 360, Foolbox, Advbox, and UnMask are also used for robustness evaluation \cite{gollner_verifai_2023}.

\begin{table}[h!]
\centering
\caption{Requirements for the second TAI principle: technical robustness and safety.}
\label{tab:Robustness}
\vspace{2mm}
\resizebox{\textwidth}{!}{%
\begin{tabular}{cm{3cm}m{6cm}ccccccc}
\toprule
\textbf{Id} & \textbf{Subprinciple} & \textbf{Requirement} & \makecell{\textbf{Priority}\\\textbf{scale}} & \makecell{\textbf{Medical}\\\textbf{process}} & \textbf{Patient} & \textbf{Clinician} & \makecell{\textbf{HC}\\\textbf{Provider}} & \makecell{\textbf{HC}\\ \textbf{Regulator}}  & \makecell{\textbf{AI}\\\textbf{developer}}\\ \midrule
R2.1 & Resilience to attack and security
& \raggedright Implement measures against data poisoning, including adversarial detection (OOD), outlier management, and specific preprocessing approaches for different data types, incorporating domain knowledge where possible
& Must
& \makecell{Model\\Development}
& I	&C, I&	C, I, M&	I&	R
\\ \midrule
R2.2 & Resilience to attack and security, Accuracy, Reliability and Reproducibility
& \raggedright The patient is provided with mechanism to withdraw consent, allowing them to stop the AI system from using their data in subsequent training sets& Must
& \makecell{Model\\Development}
& I	&C, I&	C, I, M&	I&	R
 \\ \midrule
R2.3 & Resilience to attack and security, Fallback plan and general safety
& \raggedright Provide measures to guard against hardware attacks by ensuring rapid replacement and persistence mechanisms for critical machines or servers running the primary model
& Should
& All&I	&I	&R	&I	&C, M
 \\ \midrule
R2.4 & Fallback plan and general safety
& \raggedright Offer support strategies to recover the latest stable release of the software using a version control system with clear information on commits to the main repository branch
& Must
& All& I&	I&	C, I, M&	I&	R
 \\ \midrule
R2.5 & Fallback plan and general safety
 & \raggedright Enable error detection in data inputs, including managing missing data, outliers,  and variables encoding.
 &Must
 &All & I	&U&	U, M&	I&	R
 \\ \midrule
R2.6 & Fallback plan and general safety
 & \raggedright Enable error or warning notifications and provide mechanisms for human intervention before continuing the decision-making process.
 & Should
 &All & A, I&	U&	U, M&	I&	R
 \\ \midrule
R2.7 & Accuracy, Reliability and reproducibility
 & \raggedright Incorporate training strategies to minimize overfitting and enhance generalizability, including cross-validation (K-fold CV, Nested CV, Leave-one-out CV), regularization techniques, and data augmentation, with the recommendation to integrate domain-specific knowledge where applicable.
 &Should
 &\makecell{Model\\Development}
 & I	&C, I&	C, I, M&	I&	R
\\ \midrule
R2.8 &Accuracy, Reliability and reproducibility
 & \raggedright  Include debias training by removing non-robust or meaningless features, informed by clinician inspection, to ensure that only relevant features are utilized
 &Should
 &\makecell{Model\\Development}
 &  I	&C, I&	I, M&	I&	R
 \\ \midrule
R2.9 & Accuracy, Reliability and reproducibility
& \raggedright Present performance results with confidence intervals and statistical significance indicators to enable clinicians and healthcare providers to select the best-performing models.
&Should
&All
& I	&U&	U, M&	I&	R
 \\ \midrule
R2.10 & Accuracy, Reliability and reproducibility
& \raggedright Allow relevant stakeholders, including clinicians, to agree on target performance levels (e.g., pre-set thresholds) and the appropriate metrics based on the specific clinical prediction goals and endpoints
&Must
&\makecell{Model\\Development}
& I&C, I&	C, I, M&	I&	R
\\ \midrule
R2.11 & Reliability and reproducibility
& \raggedright Use sufficient training data that reflects the heterogeneity of clinical cases, including diverse patient genotypes, phenotypes, and geographic locations
&Should
&\makecell{Model\\Development}
& I, A&	I&	C, I, M&	I&	R
 \\ \midrule
R2.12 & Reliability and reproducibility
& \raggedright Utilize health data standards and common structures, such as OMOP-CDM, to ensure consistency in data handling
&Could
&\makecell{Model\\Development}
& I	&I, U&	R, U&	M&	C
\\ \midrule
R2.13 & Accuracy, Reliability and reproducibility
& \raggedright Offer mechanisms for continuous monitoring and model improvement, such as updating with new data and regular model retraining
&Should
&All
& I	&I, C&	C, I, M&	I&	R
\\ \midrule
R2.14 & Reliability and reproducibility
& \raggedright Support open science directives by making AI model code or metadata available for other researchers
&Could
&\makecell{Model\\Development}
& I	&I&	I&	M, I& 	R
\\ \bottomrule
 \multicolumn{10}{l}{$\ast$ R: Responsible, A: Affected, U: Use, M: Monitor, C: Contribute, I: Informed, HC: Healthcare}
\end{tabular}%
}
\end{table}

\paragraph{Use Case: TAI in CVD} Ensuring technical robustness and safety is essential for the deployment of AI systems in CVD applications, where incorrect outputs may incur high clinical risks. While many AI models demonstrate impressive accuracy in controlled settings—such as detecting arrhythmias or interpreting cardiac images—translating this performance into real-world clinical environments remains a significant challenge. A recent scoping review revealed that only around 17\% of cardiovascular AI systems had been tested in randomized clinical trials, and just one-third shared their code or data for external scrutiny \cite{Moosavi2024}. This lack of rigorous evaluation raises concerns about unanticipated failure modes, particularly in diverse or underrepresented patient populations. Healthcare systems are encouraged to implement dedicated teams to track algorithmic performance post-deployment, addressing issues like data drift or performance degradation. Collaboration with cardiologists is critical throughout this process: their clinical expertise ensures that models are grounded in evidence-based predictors of CVD, while also helping identify spurious correlations that could compromise diagnostic accuracy \cite{mihan2024mitigating}. Furthermore, medical experts play a key role in defining the AI system’s operational design domain (ODD \cite{herrera2025responsible})—the specific clinical scenarios, patient populations, and data conditions under which the model is intended to function. Such expert-driven delineation enhances preparedness for out-of-distribution (OOD) data and improves the contextual robustness of AI systems, ultimately aligning their operation with the complexities of real-world clinical CVD uses.

\subsection{Privacy and data governance (data protection, data quality and access)}
\label{subsec43}

Privacy is a crucial aspect in medical AI and the broader digital health sector, as hospitals have the weakest cybersecurity among industries as shown by increasing cyberattacks in Europe and the U.S. over the last decade \cite{saw_current_2022}. Other key concerns include data shared without consent, inappropriate consent forms, data repurposing without patient knowledge, personal data exposure, and data confidentiality/privacy breach. Table \ref{tab:Privacy} presents the proposed features and requirements aimed at addressing the privacy and data governance principle.

AI systems must minimize risks of losing control over personal information, particularly for those impacted by AI decisions \cite{gupta_role_2023}. Principles of Privacy-by-Design advocate for minimizing risks by ensuring that only the necessary volume, type, and quality of data are used for specific activities \cite{oprescu_towards_2022}. Multiple strategies are available to safeguard privacy, including anonymization, which eliminates both direct identifiers that explicitly link to individuals and indirect identifiers, which can reveal identities when combined with other information. Although these suppression techniques improve privacy, they may compromise data utility. Alternatively, pseudonymization replaces personal identifiers with pseudonyms, preserving data subject characteristics while enhancing privacy protection \cite{kaur_trustworthy_2022}. Furthermore, homomorphic encryption technology has gained attention for health data security, as it allows encrypted data to be used for analysis without needing decryption \cite{kim_requirements_2023}. Centralized storage of sensitive health data increases the risk of cyberattacks, making federated learning a valuable alternative by enabling model training across multiple locations without transferring raw data \cite{kim_requirements_2023,luzon2024tutorial}. However, its server-client architecture remains susceptible to server failures, which can compromise model accuracy. Additionally, concerns about model disclosure and other risks \cite{rodriguez2023survey} may deter researchers from adopting this approach. To enhance privacy in federated learning, secure aggregation techniques such as geometric median, trimmed mean, blockchain, and Privacy Enhanced Federated Learning (PEFL) have been proposed \cite{kaur_trustworthy_2021, yang_trustworthy_2023}.

Various frameworks have been developed to assess privacy issues including Privacy Meter, IBM differential privacy toolkit, and Tensor privacy. In addition, privacy leakage can be tested using  Membership Inference Attacks (MIA) approaches tailored to the data type  \cite{gollner_verifai_2023}. Mittal et al. proposes a privacy rubric based on the presence of privacy-compromising information in the annotations \cite{mittal2024responsible}

A key challenge in medical AI is ensuring the availability and quality of data used for model development and validation, as these factors directly impact predictive performance \cite{alzubaidi_towards_2023}. Clinical data, often collected for routine care rather than research, may be biased, incomplete, or contain errors. Additionally, establishing a reliable "ground truth" requires expert labeling, introducing variability and uncertainty in model interpretation. While EHR serve as a vital data source for AI-driven analysis, their inconsistent quality hinders the development and effective implementation of AI models. To address data governance challenges in healthcare, fostering a strong global digital ecosystem can produce high-quality and representative data securely accessible to researchers \cite{hashiguchi_fulfilling_2022}.  To address these issues, robust data governance is essential for ensuring the safety and reliability of clinical models, guided by standards for data interoperability (e.g., FHIR-Fast Health Interoperability Resources-, SNOMED-Systemized Nomenclature of Medicine-) and dataset descriptions (e.g., OMOP CDM, FAIR principles-Findability, Accessibility, Interoperability, and Reusability-) that regulate data quality, completeness, and accessibility \cite{hashiguchi_fulfilling_2022}. This includes clear protocols for data storage, retention, quality control, access management, and patient confidentiality \cite{saw_current_2022, markus_role_2021}. 

Data harmonization and data governance are closely linked in medical and clinical applications, particularly in the development of TAI. Data harmonization ensures that health data from different sources, formats, and systems are standardized and made interoperable, which is essential for building reliable AI models \cite{nan2022data}. However, effective harmonization cannot occur without strong data governance. For example, when integrating EHRs from multiple hospitals, governance policies are needed to define how patient identifiers are matched, how conflicting entries are resolved, and which interoperability standards should be applied. Similarly, in multi-center clinical trials, governance ensures that data collection protocols are consistent and that privacy regulations such as Health Insurance Portability and Accountability Act (HIPAA) or GDPR are upheld across sites. Without clear data governance frameworks, harmonization efforts may result in inconsistent, biased, or incomplete datasets, ultimately undermining the transparency and reliability of AI-driven clinical decisions. Together, harmonization and governance form the backbone of TAI, ensuring that medical data is accurate, secure, and ethically usable. 

When evaluating the data collection processes, Han and Choi propose a concise 5-item checklist, covering data management, bias mitigation, outlier identification, adversarial attack prevention, and data readiness for training \cite{han2022checklist}. Similarly, the Korea National IT Industry Promotion Agency offers a self-inspection checklist for AI developers, comprising 16 items across six principles—legitimacy, safety, transparency, participation, responsibility, and fairness—to assess the impact of AI on personal data and reduce bias \cite{kim_requirements_2023}.

A promising solution tackling both privacy and data governance in healthcare is synthetic data generation, which produces data with statistical properties akin to real patient data. The validity of decisions based on synthetic data relies on the quality of the training data, specifically its representativeness of the target patient cohort, diversity to ensure robust outcomes, careful management of selection bias to address underrepresented groups, and record completeness to handle preprocessing challenges like missing data and outliers. The quality of synthetic data is assessed based on similarity to real data, usability in clinical contexts, privacy (minimizing patient re-identification risks), and fairness. These factors must be rigorously validated during both data generation and application phases to ensure safe and effective use in healthcare \cite{vallevik_can_2024}. For this, a clear and well-defined operational design domain is fundamental, as it sets the boundaries for appropriate use cases, informs the necessary characteristics of synthetic data, and helps ensure alignment between the synthetic dataset and the real-world deployment conditions of the AI tool \cite{herrera2025responsible}.
\begin{table}[h!]
\centering
\caption{Requirements for the third TAI principle: privacy and data governance.}
\label{tab:Privacy}
\vspace{2mm}
\resizebox{\textwidth}{!}{%
\begin{tabular}{cm{3cm}m{6cm}ccccccc}
\toprule
\textbf{Id} & \textbf{Subprinciple} & \textbf{Requirement} & \makecell{\textbf{Priority}\\\textbf{scale}} & \makecell{\textbf{Medical}\\\textbf{process}} & \textbf{Patient} & \textbf{Clinician} & \makecell{\textbf{HC}\\\textbf{Provider}} & \makecell{\textbf{HC}\\ \textbf{Regulator}}  & \makecell{\textbf{AI}\\\textbf{developer}}\\ \midrule

R3.1 & Privacy and data protection
& \raggedright Model training and testing is conducted in a secure and controlled environment to prevent unauthorized access
& Must
& \makecell{Model\\Development}
&I&	 I&	C, M&	I&	R
 \\ \midrule
R3.2 & Privacy and data protection
& \raggedright Privacy is continuously monitored to adjust privacy parameters throughout AI system lifecycle
& Should
&All
& A, I&	A, I&	I, M&	I	&R
\\ \midrule
R3.3 & Privacy and data protection
& \raggedright Privacy measures adopted are compliant with local and regional data protection regulations
& Must
&All
& A, I&	A, I&	A, I&	C, M&	R
 \\ \midrule
R3.4 & Privacy and data protection
 & \raggedright Regardless of its type, data used for training or testing are anonymized/pseudonymized or apply other privacy techniques (differential privacy, homomorphic encryption)
 &Must
 &\makecell{Model\\Development}
 & A, I&	A, I&	I, R&	I, M&	C
\\ \midrule
R3.5 & Privacy and data protection
 & \raggedright Only the minimum data necessary for prediction purposes is collected, adhering to the data minimization principle
 &Should
 &\makecell{Model\\Development}
 &A, I&	A, I&	C, M&	I&	R
 \\ \midrule
R3.6 &Privacy and data protection
 & \raggedright  Assess and provide contingency measures to assure that privacy mechanisms do not significantly degrade model performance
 &Would
 &\makecell{Model\\Development}
 &  A, I&	I&	I, M&	I	&R
 \\ \midrule
R3.7 & Privacy and data protection
& \raggedright Consider federated learning approaches when the model uses data located at various sites (hospitals, care centers, etc.)
&Should
&All
& I	&I&	C, M	&I&	R
 \\ \midrule
R3.8 & Privacy and data protection
& \raggedright If federated learning is implemented, privacy protection is guaranteed using secure aggregation techniques
&Should
&All
& A, I&	A, I&	C, M&	I&	R
 \\ \midrule
R3.9 & Privacy and data protection, quality and integrity of the data
 & \raggedright Restrict the use of sensitive personal information that may be subject to discrimination or lacks predictive utiliy
 & Should
 &All & A, I&	A, I&	R&	I, M&	C
 \\ \midrule
R3.10 & Quality and integrity of data
& \raggedright During data collection, avoid socially constructed biases (incorporate domain clinical knowledge)
&Should
&\makecell{Model\\Development}
& I&	C&	C, M&I&	R
 \\ \midrule
R3.11 & Quality and integrity of data
& \raggedright During data collection, avoid innacuracies and missing information (incorporate domain clinical knowledge)
&Should
&\makecell{Model\\Development}
& I	&C	&C, M	&I	&R
 \\ \midrule
R3.12 & Quality and integrity of data
& \raggedright Ensure that labeled data is based on a clinical "ground truth" 
&Should
&\makecell{Model\\Development}
& A, I&	C&	M, I	&I&	R
\\ \midrule
R3.13 & Quality and integrity of data
& \raggedright Define a data management plan with clear protocols for data collection and data storage
&Should
&All
& I&	I&	C, M	&I&	R
\\ \midrule
R3.14 & Quality and integrity of data
& \raggedright Perform an exploratory data analysis prior to training to ensure the quality of data (incorporate domain clinical knowledge)
&Should
&\makecell{Model\\Development}
& I	&C&	C, M&	I&	R
 \\ \midrule
R3.15 & Quality and integrity of data
& \raggedright Maintain a comprehensive documentation of the dataset employed, including relevant metadata
&Should
&\makecell{Model\\Development}
& I&	I&	C, M	&I&R
 \\ \midrule
R3.16 & Access to data
& \raggedright Define a protocol for data access considering access rights of  models' users  and users affected by model's decisions.
&Should
&All
& A, I&	A, U&	R&	I&	C, U
\\ \midrule
R3.17 & Quality and integrity of data, Access to data
& \raggedright Utilize health data standards and common structures (e.g. OMOP-CDM, FAIR), to ensure good data governance
&Could
&All
& I	&I, U&	R&	M&	C
\\ \bottomrule
 \multicolumn{10}{l}{$\ast$ R: Responsible, A: Affected, U: Use, M: Monitor, C: Contribute, I: Informed, HC: Healthcare}
\end{tabular}%
}
\end{table}

\paragraph{Use Case: TAI in CVD} In medical AI systems for CVD, vast amounts of sensitive health data (e.g. EHR, imaging, wearable sensor readings) fuel algorithm development. Recent studies have made significant strides in preserving patient privacy and improving data governance, by exploring controlled data sharing through anonymization and synthetic data generation. Johann et al. \cite{johann2025anonymize} applied rigorous de-identification and generative modeling to heart failure records, finding that both methods (and even their combination) protected privacy while introducing only minimal deviations in computed risk scores. Likewise, a 2025 survey by Williams et al. \cite{williams2025new} on cardiovascular imaging data reports that experts view federated learning and synthetic data as promising solutions to privacy constraints, citing improved data access and multi-institutional diversity as major benefits.
Although privacy-preserving methods such as federated learning have advanced \cite{luzon2024tutorial}, significant challenges remain in applying them effectively to medical AI systems for CVD. Differences in hospital data, such as demographics or imaging protocols, can hinder federated learning performance, while deploying secure infrastructures requires technical coordination between healthcare providers with different computational capabilities \cite{choudhury2025advancing}.

\subsection{Transparency (traceability, communication, explainability)}
\label{subsec44}

A significant challenge in digital health is the lack of transparency in AI decision-making for physicians and patients, as well as ambiguity around what constitutes a suitable explanation and how to evaluate its quality \cite{markus_role_2021}. Transparency should be addressed throughout the AI lifecycle to foster trust by enabling users to understand the AI process \cite{nasarian_designing_2024}. Additionally, transparency requirements must be adapted to different stakeholders' needs and the type of information presented should consider each user's context and preferences \cite{arrieta2020explainable}, for instance, clinicians need clear explanations of decision-making processes, while healthcare providers may prioritize cost savings \cite{upadhyay_call_2023}. Transparency should also promote collaboration between AI developers and clinicians, bridging communication gaps and refining systems with clinical insights. Thus, gray-box models, which balance between black-box and transparent models, offer high utility in healthcare and can be effectively explained if well-designed \cite{nasarian_designing_2024}.

The application of explainable AI (XAI) in health is extensive and uses various intrinsic or model-agnostic methods, such as dimensionality reduction, rule extraction, feature importance, attention mechanisms, and surrogate representations \cite{bharati_review_2023}. These methods provide insights into model logic, applicable to both global reasoning and individual predictions. There are several solutions that address explainability-by-design, but some may require domain expertise and additional data \cite{salahuddin_transparency_2022}. For example, concept learning models involve predicting high-level clinical concepts, which have been previously defined and annotated in the training set by experts, allowing clinicians to use these concepts for making the final decision. Case-based models rely on class discriminative prototypes compared with input features, or latent space interpretation, where high-dimensionality feature space is reduced to uncover salient factors of the variation learned in the data with respect to the clinical knowledge. Fuzzy rule integration with complex neural networks is also used, offering semantic interoperability that mimics human reasoning. These methods show potential for integration into clinical workflows. Additionally, preprocessing steps like handling missing values, normalization, data curation, feature engineering, managing imbalanced data, and feature selection can enhance interpretability and aid in selecting the best explainable model.

Interpretability and fidelity are essential for explainability, representing how understandable an explanation is and how accurately it reflects model behavior, respectively \cite{markus_role_2021}. Additional aspects include clarity (consistent rationale for similar instances), parsimony (concise presentation of information), completeness (sufficient detail for output computation), and soundness (the truthfulness of the model). These elements offer valuable insights into data governance, helping users understand when AI solutions are reliable and what minimum data quality requirements must be met \cite{mueller_explainability_2022}. Additionally, by enabling users to effectively understand and utilize the AI model, its usability is enhanced, maximizing its potential in clinical applications \cite{upadhyay_call_2023}. Causability, or the causal understanding of explanations, is also crucial in the medical field. XAI must align explanations with clinicians' prior knowledge to support informed decision-making. 

Although post-hoc interpretability methods are among the most used in healthcare, they are merely approximations and can create a false sense of confidence regarding model behavior \cite{salahuddin_transparency_2022}. Thus, addressing interpretability from the design phase helps avoid issues with post-hoc explanations. It is preferable to create interpretable models from the beginning rather than explaining black-box models\cite{rudin2019stop}, which can lead to serious patient risks \cite{nasarian_designing_2024}. Another challenge is the low interactivity in most of the current XAI methods, which limits flexibility in adapting explanations to users' diverse contexts and technical backgrounds. These approaches offer only static explanations, which do not allow medical experts to engage in questioning—a key element for high-quality causability. This limitation, combined with the approximate nature of post hoc methods, restricts the effectiveness of XAI in healthcare settings \cite{mueller_explainability_2022}.

Most scientific papers on medical AI systems that tackle transparency focus solely on model explainability, overlooking communication and traceability aspects. Fehr et al. report that publicly available information on medical AI lacks transparency on safety, risks, and data collection details, which are key for assessing bias \cite{fehr_trustworthy_2024}. Enforcing regulations like the EU AI Act promotes transparency and encourages adherence to TAI principles. Adopting health-oriented reporting practices, such as TRIPOD, helps address traceability, though there should be room for customization based on each use case \cite{collins2024tripod+, ahmad_responsible_2023}. Additionally, to address communication, both doctors and patients must be informed beforehand that they are interacting with a decision-making AI support tool. Communication also involves ensuring that doctors understand how patients might interpret the system's outputs and verifying health information directly with them. This approach helps alleviate communication challenges and fosters patient trust in AI \cite{bharati_review_2023} . Table \ref{tab:Transparency} outlines the features and requirements identified to fulfill this Transparency principle.

Evaluating explanations is crucial to assessing the usefulness of AI decisions \cite{markus_role_2021}. User involvement is key, influencing evaluation approaches as categorized by Doshi-Velez et al. into three levels: application-grounded, human-grounded, and functionally grounded \cite{doshi2017towards}. Most metrics used in medical AI explanation evaluation are domain-independent and often overlook expert feedback, which, despite its subjectivity, is vital for making explanations useful to clinicians \cite{pietila_when_2024}. Validation metrics involving expert input, such as the System Causability Scale (SCS) \cite{holzinger2020measuring} and Trustworthy Explainability Acceptance \cite{kaur2021trustworthy} are useful for assessing the quality of explainability methods \cite{salahuddin_transparency_2022}. Involving experts in the validation of explanations often supports their qualitative evaluation, though it is typically insufficient. Therefore, quantitative metrics are also necessary to ensure explainability methods meet the desired standards, despite the challenges of directly quantifying explanations. Domain-independent metrics used for quantitative evaluation of explanations assess a variety of aspects, including explanation robustness (likelihood that similar inputs yield similar explanations), complexity (whether only highly attributed features are truly predictive), faithfulness (whether explanations capture relevant features), randomization (difference between original and random class explanations), complexity (if explanations are based on a small number of features), expressive power (structure or form of the output of the explanation method), translucency (degree of use by the explanation method of the parameters within the model), algorithmic complexity (computational cost of the explanation method), portability (applicability across different contexts), significance (quantifies the spatial precision of the XAI focus and its overlap with the segmentation-based ground truth) or fidelity (similarity of final prediction in surrogate models) \cite{pietila_when_2024, gollner_bridging_2023,gollner_verifai_2023, cao_fuzzy_2024, stodt_novel_2023}.

The features and requirements proposed to address this principle are listed in Table \ref{tab:Transparency}.
\begin{table}[h!]
\centering
\caption{Requirements for the fourth TAI principle: transparency.}
\label{tab:Transparency}
\vspace{2mm}
\resizebox{\textwidth}{!}{%
\begin{tabular}{cm{2.5cm}m{7cm}ccccccc}
\toprule
\textbf{Id} & \textbf{Subprinciple} & \textbf{Requirement} & \makecell{\textbf{Priority}\\\textbf{scale}} & \makecell{\textbf{Medical}\\\textbf{process}} & \textbf{Patient} & \textbf{Clinician} & \makecell{\textbf{HC}\\\textbf{Provider}} & \makecell{\textbf{HC}\\ \textbf{Regulator}}  & \makecell{\textbf{AI}\\\textbf{developer}}\\ \midrule
R4.1 & Traceability
& \raggedright Follow standards and best practices (e.g., TRIPOD) for reporting model development, including data collection and labeling, to ensure traceability of all components and stages
& Should
& \makecell{Model\\development}
&I&	I, C&	I, C, M&	I&	R
\\ \midrule
R4.2 & Traceability
& \raggedright Provide mechanisms or Intellectual Property (IP) compliant documentation about model components to allow for internal and external auditability by health stakeholders affected by the AI system's decisions.
& Should
& All
& I, U&	I, U&	I, U&	I, U, M&	R
 \\ \midrule
R4.3 & Explainability
 & \raggedright Offer both global and individual explanations of model predictions to support different clinical decision-making contexts.
 &Could
 &All
 & U, C, I&	U, C, I&	U, C, M&	I&	R
 \\ \midrule
R4.4 & Explainability
& \raggedright Prioritize the use of transparent models or explainability-by-design methods, while considering the tradeoff with model performance.
& Could
&\makecell{Model\\development} & C&	C& 	I, M&	I& 	R
 \\ \midrule
R4.5 & Explainability
& \raggedright The tradeoff between model performance and interpretability is addressed adequately, considering the needs of users (clinicians, patients, HC providers) for various use cases.
&Should
&All
&A, I&	U, I&	U, I, M&	I&	R
\\ \midrule
R4.6 &Explainability
& \raggedright  Aspects such as interpretability and fidelity of the explanations are addressed in cooperation with  stakeholders affected by the AI system´s decision
&Could
&All
& A, I&	I&	 I, M&	I&	R
\\ \midrule
R4.7 & Explainability
& \raggedright Consider and evaluate  causability aspects concerning the explanations, and involve experts (clinicians and HC providers) to refine or adapt XAI outputs.
&Should
&\makecell{Model\\development}
&  C, I&	C, I	&C, M&	I&	R
\\ \midrule
R4.8 & Explainability
& \raggedright Provide interactive explanations to enhance understanding by stakeholders affected by AI system decisions.
&Could
&\makecell{Model\\development}
& A, I&	C& 	C, M&	I&	R
\\ \midrule
R4.9 & Explainability
& \raggedright The explanations provided by the AI system are easy to understand, promoting usability and effective application in clinical cases
&Should
&All
& A, I&	C, U&	U, C, M&	I&	R
 \\ \midrule
R4.10 & Explainability
& \raggedright Incorporate visualization methods in explanations to enhance understanding for stakeholders.
&Could
&All
& A, U&	U&	U, I, M&	I&	R
 \\ \midrule
R4.11 & Explainability
& \raggedright Assess the influence of preprocessing steps on the explainability of the AI model to ensure transparency.
&Chould
&\makecell{Model\\development}
& A	&U&	U, I, M&	I&	R
\\ \midrule
R4.12 & Explainability, Communication
& \raggedright Provide multiple levels of explanation for the same prediction, tailored to the interests, preferences, and technical backgrounds of different stakeholders.
& Should
& All&A, I&	U, I&	U, I, M&	I&	R
 \\ \midrule
R4.13 & Explainability, Communication
& \raggedright Involve clinicians and patients in discussions with AI developers to determine how they interpret different explanation options, to better meet user needs.
& Should
& \makecell{Model\\development}&I&	I&	I&	I&	R
\\ \midrule
R4.14 &  Communication
& \raggedright Any stakeholder affected by decision-making is informed that they are interacting with an AI system.
&Must
&All
& I, A&	I, A&	I, A&	I&	R
\\ \midrule
R4.15 & Communication
& \raggedright The AI model’s capabilities and limitations are clearly communicated to avoid misunderstandings that could negatively impact the health process outcomes. 
&Should
&All
& I, A&	I, A&	I, A&	I	&R
\\ \bottomrule
 \multicolumn{10}{l}{$\ast$ R: Responsible, A: Affected, U: Use, M: Monitor, C: Contribute, I: Informed, HC: Healthcare}
\end{tabular}%
}
\end{table}

\paragraph{Use Case: TAI in CVD} Transparent and explainable AI is critical in high-stakes fields like cardiology. Black-box models that predict heart disease risk or interpret echocardiograms without offering reasons for their outputs can undermine clinician trust \cite{Niroda2025}.Several literature reviews  \cite{moreno2024ecg,Salih2023} found that while many ECG-based data driven models and cardiac imaging AI studies boast high accuracy, only a limited subset incorporated explainability methods to clarify how results were derived. This opaqueness poses ethical and practical issues: clinicians may be reluctant to act on an AI diagnosis of, say, valvular disease if they cannot justify it to the patient or understand the rationale of the model \cite{rosenbacke2024explainable}. 
To improve transparency, scientists are developing methods to peek inside the “black box” One study applied explainable AI techniques to an ECG-based model that predicts a patient’s “heart age,” revealing that the algorithm focused on ECG features (like QRS prolongation) that clinicians recognize as age-related \cite{Hempel2025}. In another study, XAI is applied to heart failure prognosis models to balance accuracy with interpretability \cite{MorenoSanchez2023} This alignment with medical knowledge both validated the model’s behavior and made its predictions more trustable. 

However, challenges remain—there is not yet a consensus on how to quantify a “good” explanation in medicine and researchers advocate for developing standardized XAI evaluation metrics and hybrid models that balance interpretability with accuracy \cite{marey2024explainability}. Without rigorous evaluation, there is a risk that “explanations” are misleading or not clinically useful. Many popular XAI methods are post-hoc approximations of complex models, so their faithfulness is not guaranteed, which can create a false sense of understanding \cite{Salih2024}.
Greater clinician involvement in designing and testing explanation methods is recommended to ensure that transparency efforts genuinely aid medical decision-making. Additionally, educating cardiologists in AI literacy would make them to interpret AI outputs and communicate them to patients more effectively.

\subsection{Non-discrimination and fairness }
\label{subsec45}

In healthcare, a decision may be deemed unfair if patients with similar health characteristics are treated differently based on protected or sensitive features (SF), such as gender, disability, or educational background, which should not influence medical outcomes \cite{el-sappagh_trustworthy_2023}. Fairness in medical AI can be classified into \emph{group} fairness (treating different patient groups similarly without considering protected features), \emph{individual} fairness (ensuring similar patients receive similar treatment), and \emph{counterfactual} fairness (ensuring a model’s decision remains unchanged even if a protected attribute is altered, implying that sensitive variables do not influence outcomes) \cite{kim_requirements_2023}. However, simply removing protected features is insufficient, as correlations with other features can still perpetuate unfairness. The relevant features and requirements associated with this principle are summarized in Table \ref{tab:Fairness}.

A major issue in AI systems is that training data often do not represent the true population distribution, due to factors like insufficient sample size, under-representation of minority groups, or regional biases \cite{hashiguchi_fulfilling_2022}. Missing values or unrecognized patients further contribute to biased predictions, which can have serious health implications \cite{dlugatch_trustworthy_2023}. These data issues, collectively termed as data collection bias, represent a model-agnostic fairness problem affecting medical AI.  For example, algorithms trained on EHRs risk inheriting flaws such as only representing individuals with access to electronic care or replicating clinician errors, including stereotypes, historical biases, or mistakes \cite{el-sappagh_trustworthy_2023, crigger_trustworthy_2022}. Therefore, the data collection process must be transparent and well-documented, as training on biased data can lead to unfair models and biased explanations, potentially compromising patient care. Unfair data can be categorized as class imbalance, such as labeling bias or the intentional underrepresentation of certain patient groups, and feature imbalance, where protected attributes are unevenly distributed across patient populations. However, in the medical domain, data imbalances often reflect real-world differences in disease prevalence, rather than a deliberate bias \cite{mittal2024responsible}. Nonetheless, fairness issues extend beyond imbalanced datasets, as unfairness can also arise from poorly defined proxies, algorithm choices, model delivery, or user interactions.

To address fairness during data collection, exploratory data analysis is crucial to understand distributions across variables and assess the balance of represented groups. This early analysis helps identify issues like minority intentional underrepresentation, which can lead to biased models \cite{ahmad_responsible_2023}. Involving domain experts in data collection and conducting statistical demographic analysis can improve fairness, although it may not entirely eliminate bias \cite{kim_requirements_2023}. Fairness techniques often focus on SF and privileged/unprivileged groups—groups more or less likely to receive positive classifications based on SF. Common approaches include No Disparate Treatment (prohibiting SFs in decisions), No Disparate Impact (equal positive classification likelihood across groups), No Disparate Mistreatment (equal misclassification rates), and Min-Max Fairness (minimizing the highest error rate among groups) \cite{li_fairness_2024}.

To address data bias, models can incorporate two fairness strategies: (1) in-processing, which modifies models during training to enhance fairness, and (2) post-processing, which corrects outputs after training. Both strategies require SFs to ensure fair outcomes, and some methods combine these by training separate classifiers for different SF groups \cite{el-sappagh_trustworthy_2023}. In-processing techniques often use regularization to penalize correlations between SFs and predicted labels, balancing performance and fairness. Common in-processing techniques include adversarial learning \cite{beutel2017data}, bandits \cite{gillen2018online}, constraint optimization \cite{celis2019classification}, regularization \cite{di2020counterfactual} , multitask modeling \cite{benton2017multi}, multitask adversarial learning \cite{beutel2017data}, and reweighting \cite{krasanakis2018adaptive}. Post-processing methods include calibration \cite{liu2017calibrated}, constraint optimization \cite{kim2018fairness}, thresholding \cite{iosifidis2019fae}, and transformation \cite{chiappa2019path}. Additionally, other methods for ensuring less biased data include adversarial learning \cite{feng2019learning}, causal methods \cite{salimi2019data}, relabeling and perturbation \cite{cowgill2017algorithmic}, (re)sampling \cite{adler2018auditing}, reweighting \cite{kamiran2012data}, transformation \cite{calmon2017optimized}, variable blinding \cite{chouldechova2017fair}, data augmentation, or stratified models by SFs \cite{szabo_clinicians_2022}.

The ``design for all'' subprinciple emphasizes inclusive design in healthcare, aiming to create systems that match the cognitive abilities of clinicians and patients \cite{xie2024towards}. Central to this approach is user-centered design, which seeks to minimize cognitive load and improve communication clarity. When applying TAI principles in healthcare, it is crucial to integrate cognitive and usability engineering to develop AI systems that align with users' needs. This ensures that AI tools are accessible, intuitive, and safe for clinicians and patients, thereby enhancing decision-making processes, improving patient outcomes, and supporting clinical workflows. In the set of requirements proposed in this framework, this subprinciple is addressed by understanding fairness and non-discrimination not only as a matter of data representativeness and algorithmic equity, but also in terms of equitable access to, and effective use of, AI tools. By considering cognitive and accessibility demands across all TAI principles, these systems can better support the diverse needs and expectations of all stakeholders in the healthcare environment.

To evaluate fairness, it is essential to focus on datasets rather than models, measuring aspects of data curation quality. However, a notable challenge arises where including sensitive attributes to enhance fairness may inadvertently compromise privacy \cite{mittal2024responsible}. Fairness assessments can be classified into four main types \cite{el-sappagh_trustworthy_2023}: similarity-based fairness, individual Fairness, global fairness, and causal fairness.  Similarity-based includes fairness through unawareness (excluding sensitive features from decision-making), fairness through awareness (providing similar outcomes for similar individuals using similarity metrics), and causal discrimination (ensuring identical outcomes for individuals differing only by sensitive features). Individual fairness requires similar cases to have similar outcomes, which can be measured using metrics like Between Group Entropy Error \cite{speicher2018unified}. Group fairness assesses model performance at the group level, using metrics such as statistical parity \cite{corbett2017algorithmic}, disparate impact   \cite{feldman2015certifying}, equalized opportunity\cite{hardt2016equality}, equalized odds, and overall accuracy equality metric \cite{berk2021fairness}, ensuring classification outcomes are independent of SFs. Lastly, causal fairness consider the outcome for every single individual by comparing the probabilities of possible outcomes for different interventions \cite{chiappa2019path}.

\begin{table}[h!]
\centering
\caption{Requirements for the fifth TAI principle: Non-discrimation and fairness.}
\label{tab:Fairness}
\vspace{2mm}
\resizebox{\textwidth}{!}{%
\begin{tabular}{cm{2cm}m{7cm}ccccccc}
\toprule
\textbf{Id} & \textbf{Subprinciple} & \textbf{Requirement} & \makecell{\textbf{Priority}\\\textbf{scale}} & \makecell{\textbf{Medical}\\\textbf{process}} & \textbf{Patient} & \textbf{Clinician} & \makecell{\textbf{HC}\\\textbf{Provider}} & \makecell{\textbf{HC}\\ \textbf{Regulator}}  & \makecell{\textbf{AI}\\\textbf{developer}}\\ \midrule
R5.1 & Fairness
& \raggedright The sensitive or protected features are identified by clinical experts
& Must
& \makecell{Model\\development}
&A, I&	R, C&	I&	I&	C, M
\\ \midrule
R5.2 & Fairness
& \raggedright Before model training, an initial exploration and pre-assessment of the collected data is conducted to detect the existence of identifiable and discriminatory bias.
& Should
& \makecell{Model\\development}
& I	&C	&C, M&	I&	R
\\ \midrule
R5.3 & Fairness
& \raggedright Develop two model versions—one including sensitive features and one without—to assess the impact of these variables on performance.
&Could
&\makecell{Model\\development}
& I&	I&	I, M&	I&	R
\\ \midrule
R5.4 & Fairness
& \raggedright Inspect the influence of protected variables on model´s decision (incorporate clinical domain knowledge)
& Should
&\makecell{Model\\development} & I&	C&	C, M&	I&	R
\\ \midrule
R5.5 & Fairness
& \raggedright During data collection, ensure the real population distribution is reflected to prevent underrepresentation of minority groups, (incorporate clinical domain knowledge)
&Should
&\makecell{Model\\development}
& A, I&	C,I&	C, R&	I&	C, M
\\ \midrule
R5.6 &Fairness
& \raggedright  Conduct statistical demographic analyses on collected data to assess whether population distribution aligns with the intended cohort
&Should
&\makecell{Model\\development}
& I&	I&	I, M&	I&	R
 \\ \midrule
R5.7 & Fairness
& \raggedright Address missing data using appropriate sampling techniques, except when it is not missing at random, while maintaining alignment with real population distribution (incorporate clinical domain knowledge)
&Should
&\makecell{Model\\development}
& I&	C&	C, M	&I	&R
\\ \midrule
R5.8 & Fairness
& \raggedright Correct, if possible, labeling mistakes or unrecognized patients in training and test datasets
&Should
&\makecell{Model\\development}
& I&	C&	C, M&	I&	R
\\ \midrule
R5.9 & Fairness
& \raggedright Fully and transparently document the data collection process, following specific healthcare data standards where applicable (e.g., HL7, FHIR, DICOM, SNOMED).
&Should
&\makecell{Model\\development}
& I&	I, C&	C, R&	I	&C, M
 \\ \midrule
R5.10 & Fairness
& \raggedright Employ in-processing techniques (adversarial learning, regularization, reweighting, etc.) during model training. Incorporate clinical domain knowledge to validate reduced-bias results
&Could
&\makecell{Model\\development}
& I	&I, C&	I, C&	I&	R
\\ \midrule
R5.11 & Fairness
& \raggedright Employ post-processing techniques bias mitigation techniques (calibration, constraint optimization, etc.).  Incorporate clinical domain knowledge to validate reduced-bias results
&Could
&All
&I	&I, C&	I, C&	I&	R
\\ \midrule
R5.12 & Universal\newline Design
& \raggedright Interaction between users and the AI system is designed to minimize cognitive load
& Should
& All& A, I	&I, C&	C, M	&I	&R
\\ \midrule

R5.13 & Universal\newline  Design
& \raggedright The AI system provides clear communication approaches tailored to its users.
& Should
& All& U, A	&U, A&	I, M&	I	&R
\\ \midrule

R5.14 &  Universal\newline  Design
& \raggedright Design the AI system interface in compliance with established accessibility and usability standards (WACG, EAA, Section 508, ARIA)
&Should
&\makecell{Model\\development}
& U, A	&U, A&	I, M&	I	&R
\\ \bottomrule
 \multicolumn{10}{l}{$\ast$ R: Responsible, A: Affected, U: Use, M: Monitor, C: Contribute, I: Informed, HC: Healthcare}
\end{tabular}%
}
\end{table}

\paragraph{Use Case: TAI in CVD} AI systems developed for cardiovascular disease must ensure equitable performance across diverse patient populations. A recent study emphasizes that fairness should be embedded throughout the AI development pipeline, from data collection and model design to validation, with the dual aim of preventing amplification of existing health disparities and actively promoting equity \cite{mihan2024mitigating}. Despite growing awareness, operationalizing fairness remains difficult. Minority populations, such as certain ethnic groups or individuals in rural areas, are often underrepresented in cardiovascular datasets, which limits the model’s ability to generalize across these groups. Kaur et al. (2024) investigated a deep learning ECG model for heart failure prognosis and discovered it performed significantly worse for young Black women than for other subgroups \cite{kaur2024race}. Moreover, CVD diagnosis typically involves multimodal data sources such as EHRs, ECG signals, echocardiograms, and advanced imaging (e.g., MRIs), many of which are not equally accessible across healthcare settings, posing a risk of excluding patients who lack access to such technologies \cite{salvi2024multi}.

To address these challenges, researchers advocate for training datasets that mirror the real-world diversity of CVD patients, encompassing a wide range of ages, sexes, ethnicity, and socioeconomic backgrounds \cite{mihan2024mitigating}. For example, one study showed that race-specific models enhanced heart failure risk prediction for Black patients by capturing distinct risk factor profiles \cite{cheema2022augmented}. AI system should also be designed to function effectively with features and data sources that are widely available, ensuring usability across different clinical contexts. This suggests that algorithmic fairness interventions may range from dataset and model redesign to deployment-time adjustments. Other works echo that biases can enter at many stages. Mihan et al. (2024) note that AI bias in CVD can arise in data collection, model training, or validation, and they present an ``AI equity framework'' covering mitigation steps throughout the AI lifecycle \cite{mihan2024artificial} Additionally, fostering diversity within development teams can improve the identification of latent biases and introduce critical domain knowledge regarding health disparities, further supporting fairness in medical AI.

\subsection{Sustainability}
\label{subsec47}

AI systems are seen as as social–technical–ecological systems as their outcomes cannot be attributed solely to technology, people or data but other complex interaction between technical components (e.g., data models and processes) and their environmental impacts \cite{romano_meanings_2021}. The TAI principle of sustainability tackles the development of environmentally friendly AI based on sustainable data sources, power supplies, and carbon-efficient training methods \cite{meiser2024survey}.

Sustainability in the domain of medical AI is crucial, particularly given the substantial energy consumption associated with powerful models that process large, complex datasets, such as biosignals, medical images, and genomic information. This energy demand affects both hospitals and data centers, contributing to significant economic costs. Consequently, a critical balance must be struck: while aiming to reduce the carbon footprint and improve the sustainability of these models, there is a risk of compromising the accuracy required for reliable diagnostic and prognostic decisions.Additionally, the storage of this large medical datasets demands significant energy resources, contributing substantially to the carbon footprint of AI in healthcare. To address this, the adoption of green data centers powered by renewable energy sources offers a promising solution, enabling the continued use of high-volume clinical data for AI training and inference while minimizing environmental impact \cite{samuel2022environmental}.

On the hardware side, the manufacturing and use of specialized AI chips carry a significant environmental footprint -- from the mining of rare materials to high energy consumption -- and contribute to electronic waste (e-waste) accumulation \cite{Ueda2024}. To combat this, researchers advocate for eco-design principles such as modular, upgradeable device architectures that extend hardware lifespans and facilitate easier component replacement and recycling. On the software side, sustainable AI development focuses on maximizing efficiency and collaboration. Promoting open-source, transparent development of medical AI models is seen as a way to avoid redundant computational effort, by sharing code and pretrained models, researchers can build on existing work rather than repeat costly training from scratch \cite{Alzoubi2024}.

Sustainability should be measured throughout the entire AI system lifecycle \cite{ligozat2022unraveling}. In the design phase, specific aspects must be considered to align models with Green AI principles, such as decisions regarding data quantization, augmentation, hyperparameter tuning, feature selection, and regularization to reduce energy consumption \cite{sikand2023green, wenninger2022sustainable}. Factors like data center location, energy mix, and hardware type also impact sustainability during deployment and inference phases. Additionally, other TAI principles, such as explainability, can enhance sustainability by revealing key features that reduce model complexity and energy use.

The requirements concerning sustainability that AI developers should consider and their relation with the healthcare stakeholders are shown in Table \ref{tab:Sustainability}.
\begin{table}[h!]
\centering
\caption{Requirements for the sixth TAI principle: Sustainability.}
\label{tab:Sustainability}\vspace{2mm}
\resizebox{\textwidth}{!}{%
\begin{tabular}{cm{2.5cm}m{7cm}ccccccc}
\toprule
\textbf{Id} & \textbf{Subprinciple} & \textbf{Requirement} & \makecell{\textbf{Priority}\\\textbf{scale}} & \makecell{\textbf{Medical}\\\textbf{process}} & \textbf{Patient} & \textbf{Clinician} & \makecell{\textbf{HC}\\\textbf{Provider}} & \makecell{\textbf{HC}\\ \textbf{Regulator}}  & \makecell{\textbf{AI}\\\textbf{developer}}\\ \midrule
R6.1 & Sustainable and environmentally friendly AI
& \raggedright Efficiency parameters associated with the AI solution are measured at all stages of the solution lifecycle
& Should
& All
&I	&I&	A, M, U&	I&	R
\\ \midrule
R6.2 & Sustainable and environmentally friendly AI
& \raggedright  Models with lower complexity that maintain satisfactory accuracy are prioritized to minimize carbon footprint.
& Should
& \makecell{Model\\development}
&I&	I&	A, M, C&	I&	R
\\ \midrule
R6.3 & Sustainable and environmentally friendly AI
& \raggedright If possible, the use of pretrained models and transfer learning is promoted to enhance energy efficiency.
& Could
& \makecell{Model\\development}
&I	&I	&A, M&	I&	R
\\ \midrule
R6.4 & Sustainable and environmentally friendly AI
& \raggedright Hardware and data centers used comply with energy and resource efficiency certifications
& Could
& All
&I&	I&	R, U&	I, M&	R
\\ \midrule
R6.5 & Sustainable and environmentally friendly AI
& \raggedright During data collection, aspects such as data augmentation and quantization techniques are considered to reduce energy consumption.
& Could
& \makecell{Model\\development}
&I	&I	&A, M, C	&I&	R
\\ \midrule
R6.6 & Sustainable and environmentally friendly AI
& \raggedright During training model aspects such as hyperparameter tuning techniques, feature selection or regularization methods are considered to reduce de energy consumption.
& Should
& \makecell{Model\\development}
&I&	I&	A, M&	I	&R
\\ \midrule
R6.7 & Sustainable and environmentally friendly AI
& \raggedright Factors such as energy mix, data center location and hardware used during AI solutions deployment are considered
& Could
& All
&I&	I&	R, U&	I&	R, M
\\ \bottomrule
 \multicolumn{10}{l}{$\ast$ R: Responsible, A: Affected, U: Use, M: Monitor, C: Contribute, I: Informed, HC: Healthcare}
\end{tabular}%
}
\end{table}

\paragraph{Use Case: TAI in CVD} Medical AI systems should not only contribute to improve CVD medical processes but also minimize environmental harm. The development and deployment of AI are energy-intensive – training complex deep learning models on cardiac imaging or genomic data can consume vast computational resources, translating to a sizable carbon footprint. Paradoxically, an AI designed to save lives (through better cardiac risk prediction) might indirectly impact health via climate change if its energy use is unchecked \cite{Coghlan2023}.

To operationalize this principle, experts call for a broadening of stakeholders and metrics when evaluating AI in healthcare. Instead of focusing narrowly on accuracy or cost-effectiveness, developers are urged to factor in carbon emissions, electronic waste, and community impact. For example, an AI algorithm for cardiac imaging might be optimized not only for speed but also for energy efficiency, or run on hardware powered by renewable energy. Additionally, given the active research focus on CVD in the AI community, there is a wealth of pre-existing, well-evaluated models available. Leveraging these models through transfer learning significantly reduces the need for training from scratch, thereby decreasing energy consumption and supporting more sustainable AI development. This approach not only minimizes environmental impact but also accelerates the deployment of effective and efficient AI solutions in the clinical management of CVD.

\subsection{Auditability and accountability}
\label{subsec46}

Healthcare often struggles with the adoption of digital innovations, and medical AI systems are no exception, with many never advancing beyond the preclinical or pilot stages \cite{procter_holding_2023}. Identifying who is responsible for decisions in healthcare is crucial; however, when AI models are used in clinical decision-making, a “responsibility gap” can emerge, making it unclear whether clinicians or the AI system should be held accountable for decisions made with the assistance of such tools \cite{duran2021afraid}.

To address this challenge, the development and deployment of Responsible AI systems in healthcare is essential. A Responsible AI system requires ensuring auditability and accountability throughout its design, development, and use, in accordance with the specifications and applicable regulations of the domain in which it is deployed \cite{diaz2023connecting}. On one hand, auditability serves as a foundational component to support accountability. It involves thorough validation of the AI system’s conformity with vertical (sector-specific) regulations, horizontal (AI-wide) regulations such as the EU AI Act, and the constraints imposed by its intended application. Auditability encompasses properties such as transparency (e.g., explainability and traceability) and technical robustness, though it may not cover all dimensions of TAI—focusing instead on those dictated by ethics, regulation, and domain-specific testing protocols. Accountability, on the other hand, refers to the assignment of liability for decisions informed by the AI system's outputs, assuming prior compliance has been confirmed through auditing. It reflects the degree to which stakeholders are held responsible based on the specific regulatory and ethical frameworks governing the healthcare setting.

The accountability of AI systems in healthcare is crucial due to the significant impact that incorrect AI-driven decisions can have on patients' lives. However, determining accountability is complex, as it is distributed among multiple stakeholders, with different expertise, professional standards, and goals \cite{crigger_trustworthy_2022}. Therefore, TAI will likely need various forms of accountability, tailored to the needs of specific users, settings, timings, and circumstances  \cite{procter_holding_2023}. For instance, algorithmic accountability applied to medical AI system involves evaluating algorithms and determining who are responsible for any harm caused to those stakeholders affected directly or indirectly by the system´s decision. Assigning different priorities to system´s functions can help resolve conflicts in determining accountability in healthcare \cite{kaur_trustworthy_2022}. 

Additionally, auditability introduces key concepts such as verification, which ensures that system specifications are met, and validation, which compares the system to stakeholders' needs. Auditing AI operations in healthcare is essential not only to comply with legal requirements but also to build trustworthy, socially beneficial, and responsible AI systems. For instance, the EU AI Act specifies an audit process that includes conformity assessments, specifically for domains tagged as high-risk such as healthcare, in areas such as data governance, documentation, transparency, human oversight, robustness, accuracy, and security. Explainability is crucial in this context, entangling with other TAI principle of transparency, as it facilitates auditors' and developers' access to the model's logic and outputs \cite{herrera2025responsible}. An effective audit process might include steps like documentation review, code review, reproduction of model training, testing, performance measures, perturbation tests, and the development of alternative models. The audit report should provide a comprehensive description of the AI system's artifacts, including AI opportunities versus risks, risk management, AI model reporting methodology, audit process, quality assurance, data properties, algorithm design, and assessment metrics \cite{schwarz2024designing}. Additionally, beyond the legal implications of auditing AI systems, clinicians' decision-making performance can change over time, necessitating ongoing monitoring and periodic audits. From a technological perspective, data and concept drift, where an AI system's performance changes over time, also highlight the need for monitoring and auditing procedures to detect changes that could jeopardize patient safety.

In addition to assigning responsibility, it is crucial to implement remedies such as safe mode operation and error reporting when harm occurs. These measures ensure immediate safety and transparency \cite{han2022checklist}. Redress mechanisms are also essential, allowing stakeholders to step away from AI-assisted workflows and reinforce human agency, which must be guaranteed in any medical AI system. These mechanisms provide a way to address and correct issues, ensuring that human oversight remains a central component of healthcare decision-making \cite{fanni2023enhancing}. 

When assessing the accountability of AI system decisions, several approaches can be taken. First, conduct an algorithmic impact assessment to identify potential risks and harms when stakeholders interact with the AI, and link these to an accountability governance framework involving various stakeholders. Next, establish an accountability relationship between stakeholders interacting with the AI system and the defined impacts. This relationship should be evaluated by other actors, such as the legal services of healthcare providers, who can propose necessary changes and enforce consequences based on the system's impact \cite{metcalf_algorithmic_2021}.

The requirements concerning accountability and auditability that AI developers should be considered and their relation with the healthcare stakeholders are shown in Table \ref{tab:Accountability}.
\begin{table}[h!]
\centering
\caption{Requirements for the seventh TAI principle: accountability.}
\label{tab:Accountability}
\vspace{2mm}
\resizebox{\textwidth}{!}{%
\begin{tabular}{cm{2cm}m{7cm}ccccccc}
\toprule
\textbf{Id} & \textbf{Subprinciple} & \textbf{Requirement} & \makecell{\textbf{Priority}\\\textbf{scale}} & \makecell{\textbf{Medical}\\\textbf{process}} & \textbf{Patient} & \textbf{Clinician} & \makecell{\textbf{HC}\\\textbf{Provider}} & \makecell{\textbf{HC}\\ \textbf{Regulator}}  & \makecell{\textbf{AI}\\\textbf{developer}}\\ \midrule
R7.1 & Accountability
& \raggedright After system deployment, document how any ethical concerns have been mitigated or corrected.
& Should
& All
&A, I&	A, I&	C&	M&	R
\\ \midrule
R7.2 & Accountability
& \raggedright  Clearly define the roles and responsibilities of developers and stakeholders.
& Must
& \makecell{Model\\development}
&C,I&	C, I&	C, I, M	&I&R
\\ \midrule
R7.3 & Accountability
& \raggedright Assess the potential impacts (economical, ethical, societal, health, etc.) produced by  the AI system's functionalities, considering all stakeholders affected directly or indirectly
& Should
& \makecell{Model\\development}
&C,I	&C, I&	C, I, M&	C, I&	R, C
\\ \midrule
R7.4 & Auditability
& \raggedright Assess AI system compliance with current regulations and policies (e.g., AI Act, EU GDPR)
& Should
& \makecell{Model\\development}
&I&	I&	I, M&	I&	R
\\ \midrule
R7.5 & Auditability
& \raggedright Document any changes in clinicians' decision-making processes that could affect the AI system's functioning.
& Should 
& All
&I&	R, C&	R, C&	I&	M, C

\\ \midrule
R7.6 & Auditability
& \raggedright Report any data and concept drift in the AI system that could impact clinicians' decision-making processes.
& Should
& All
&A, I	&A,I&	I, M	&I&	R, C
\\ \midrule
R7.7 & Redress
& \raggedright Implement remedies , error reportings and compensation approaches that empower an active human agency when harm occurs
& Should
& All
&I, U	&I, U&	I, U, M&	I&	R
\\ \bottomrule
 \multicolumn{10}{l}{$\ast$ R: Responsible, A: Affected, U: Use, M: Monitor, C: Contribute, I: Informed, HC: Healthcare}
\end{tabular}%
}
\end{table}

\paragraph{Use Case: TAI in CVD} Clear accountability for AI-driven decisions in cardiovascular care is vital to maintain trust and uphold ethical standards. With multiple stakeholders (engineers, clinicians, etc.) involved in an AI system’s lifecycle, determining who is answerable for an erroneous recommendation can be difficult, which also hinder its adoption. Thus, clinicians may fear legal risks from using opaque models, while developers worry about liability beyond their control. To address this, recent work \cite{Niroda2025} emphasizes the need for accountability frameworks, where stakeholders in CVD care, alongside AI developers, must identify those affected by system decisions, prioritizing those at highest risk, and align accountability with the potential harm AI decisions may cause \cite{szabo_clinicians_2022}. These measures aim to make AI systems in CVD care more transparent and scrutable, so that when errors occur, one can trace their origin (data issues, model faults, or misuse) and assign responsibility accordingly. In practice, this could mean ensuring cardiologists retain ultimate responsibility and holding developers to safety and efficacy standards under regulatory oversight.

The auditing process for medical AI systems designed for CVD should adhere to the same principles applied to other medical AI systems, including conformity assessments to comply with regulatory frameworks such as the AI Act. Furthermore, maintaining comprehensive records of data usage, model development, and updates is essential to uphold accountability and facilitate effective auditing practices. On the clinical side, a key recommendation is to improve the engagement and training of all stakeholders with AI. As Mooghali et al. found out, involving physicians, developers, and even patients in the AI implementation process builds mutual understanding of the technology’s limits and ensures someone is clearly ``on point'' for each stage of AI use \cite{Mooghali2024}.
\section{Tradeoffs between TAI principles and requirements}
\label{sec5}

The overarching goal of this work is to create a common framework for AI developers and other stakeholders to assess whether the AI tools they are considering developing or implementing align with TAI principles. However, this initiative is far from being a one-size-fits-all solution. The particularities and intricacies of each medical AI system—such as the scope of the systems, interactions with different stakeholders, various data modalities, and the traits and boundaries of each medical domain—make it challenging to apply a universal approach. Consequently, potential tradeoffs may arise between the requirements detailed in Section \ref{sec4}, as different perspectives or intentions regarding what stakeholders aim to achieve or receive from the system can conflict. In this section we enumerate and describe various tradeoffs identified during our research, elaborating on their rationale (\emph{Why?}) and proposing potential paths to address them effectively (\emph{How?}).
 
\subsection{Technical Robustness versus Fairness}\label{subsubsec511}

The healthcare domain presents unique challenges where both technical robustness and fairness are critical. While robustness—often associated with high accuracy—is essential for producing precise diagnoses and effective treatment plans, fairness ensures that all patient groups receive equitable care. However, the under-representation of certain groups (particularly prevalent in healthcare) can lead to biased outcomes, even when the model demonstrates high accuracy overall. AI systems often perform well on majority groups but fail to generalize for underrepresented populations, resulting in disparities in predictive outcomes.

To mitigate these biases, adjustments may be necessary, even at the cost of a slight reduction in overall accuracy. Bias mitigation strategies, such as regularization techniques, reweighting, and adversarial debiasing, can help ensure more equitable treatment across different demographic groups. However, these measures must be carefully evaluated through a multidisciplinary approach, involving diverse stakeholders and exploratory use cases. This assessment should determine whether the tradeoffs in accuracy are ethically and clinically acceptable, ensuring that AI-driven medical decisions align with professional deontological standards.

\subsection{Accuracy versus Transparency}\label{subsubsec512}

Despite a potential tradeoff with model performance, interpretability has become a key focus in medical AI applications, ensuring that results are easily understandable for patients and clinicians. Increasingly, clinicians recognize the importance of technological transparency in fostering trust and facilitating clinical adoption \cite{wang_accelerating_2023}. However, complex models that achieve high accuracy often function as black boxes, obscuring their decision-making logic from users \cite{arrieta2020explainable}.

To address this, clinicians should define the target accuracy that AI systems must achieve to ensure both performance improvements and patient safety. This target should fall within a confidence interval that balances explainability and accuracy, preventing performance degradation that could compromise clinical outcomes. Hybrid approaches, such as employing interpretable models for initial decision-making and complex models for confirmatory analysis, may offer a viable solution. Ultimately, the level of required explainability depends on both the stakeholder and the healthcare context. Clinicians may need detailed insights to validate AI recommendations, while patients may require simplified explanations to understand their diagnosis or treatment options. In high-risk, life-threatening conditions, maximizing accuracy may take precedence over full explainability, whereas in routine diagnostics, interpretability might be prioritized to enhance transparency and trust.

\subsection{Fairness versus Data Governance}\label{subsubsec513}

Ensuring fairness in AI-driven healthcare systems requires high-quality, representative data, yet this priority can clash with strict data governance policies. While governance frameworks regulate how patient data is collected, stored, and shared to ensure compliance with privacy laws (e.g., GDPR, HIPAA), they may also restrict access to diverse datasets, potentially leading to biased AI models. Limited access to sensitive attributes—such as ethnicity, gender, or socioeconomic status—can hinder fairness assessments, making it difficult to identify and mitigate disparities. If AI models are trained on incomplete or non-representative datasets, they may perform well for majority populations but fail to generalize across underrepresented or marginalized groups, reinforcing existing healthcare disparities.

A potential solution to address data access limitations is the use of synthetic data. However, synthetic datasets must be rigorously validated to ensure they accurately represent diverse populations and do not introduce bias artifacts that could compromise fairness. To achieve this, stakeholders must be actively involved in delineating the operational design domain of the AI system, ensuring that synthetic data generation aligns with realistic clinical conditions and intended use cases. Moreover, many fairness-enhancing techniques, such as reweighting, adversarial debiasing, or domain adaptation, require access to protected attributes to detect disparities. However, governance policies may prohibit their use. To achieve both fairness and robust data governance, collaboration between AI developers, ethicists, and policymakers is essential, ensuring that governance frameworks support bias mitigation efforts while upholding patient privacy and security.

\subsection{Fairness versus Privacy}\label{subsubsec514}

Many healthcare datasets contain sensitive attribute information, which, while essential for fairness analysis, also increases the risk of privacy breaches. This creates a fairness–privacy paradox, where enhancing one aspect compromises the other. Sensitive features help models mitigate biases and improve fairness in decision-making. One potential solution is to provide aggregated population statistics instead of individual-sensitive attribute labels in publicly available datasets. However, most fairness algorithms rely on these annotations for evaluation, and their absence may hinder bias detection and mitigation \cite{mittal2024responsible}.

To balance this tradeoff without losing the nuanced insights that sensitive attributes provide for bias mitigation, privacy-preserving techniques such as encryption or anonymization could be implemented. However, these methods risk distorting data distributions, potentially amplifying biases rather than reducing them, thereby complicating fairness assessment and correction. The regulatory framework governing AI systems, such as GDPR and HIPAA, plays a critical role in determining how and when sensitive attributes can be utilized. Ultimately, the balance between privacy and fairness depends on the specific use case. In population-level studies, privacy may take precedence, whereas in individualized AI-driven treatment decisions, fairness must be safeguarded to prevent disparities in patient care.

\subsection{Accuracy versus Privacy}\label{subsubsec515}

Ensuring patient privacy while maintaining high model accuracy is another significant challenge in medical AI. In applications such as predictive diagnostics or individualized treatment plans, high accuracy is essential. However, privacy measures that distort data distributions can lead to misdiagnoses or suboptimal treatment recommendations. 

To address this tradeoff, AI developers and healthcare stakeholders must work together to implement privacy-preserving methods that minimize accuracy loss while ensuring robust patient data protection. Privacy-enhancing techniques, including differential privacy, homomorphic encryption, and federated learning, are designed to protect sensitive health data but often come at the expense of model performance. Differential privacy, for instance, introduces controlled noise into the data to obscure individual identities. However, excessive noise can degrade accuracy and fairness, disproportionately affecting underrepresented groups with already limited data representation. Homomorphic encryption allows computations on encrypted data, preventing direct access to sensitive patient information, but requires substantial computational resources, potentially slowing down real-time decision-making in critical scenarios such as emergency medicine. Similarly, federated learning enables multiple healthcare institutions to collaboratively train AI models without sharing raw patient data, yet issues like model leakage, communication overhead, and lack of data standardization can reduce model effectiveness.
 
\subsection{Privacy versus Transparency}\label{subsubsec516}

Privacy and transparency are both fundamental principles of TAI, but they often come into conflict in medical AI applications. Achieving transparency often necessitates revealing details about the data, model parameters, or training processes, which can inadvertently compromise patient privacy. Therefore, a challenge arises from the need for providing meaningful explanations regarding AI decisions without exposing sensitive patient information. XAI techniques generate human-interpretable insights into model predictions, but if explanations include specific patient data or feature importance rankings that reveal personal health details, privacy risks may increase. Another aspect of this tradeoff emerges from federated learning. Although this approach prevents the direct exchange of patient records, it reduces transparency, as stakeholders have limited visibility into how data from different sources contribute to model outcomes. This opacity may hinder trust among clinicians and patients, raising concerns about AI-driven medical decisions, as well as their accountability.

To navigate this tradeoff, AI developers and healthcare stakeholders must adopt privacy-preserving transparency strategies, such as privacy-aware explanation techniques, or aggregated feature importance reports. A risk-based approach can help determine the appropriate level of transparency based on the clinical context, ensuring that AI systems remain both interpretable and compliant with patient privacy standards.

\subsection{Sustainability versus Transparency} \label{subsubsec517}

Ensuring transparency in AI models often involves complex post-hoc explainability methods, which may significantly increase computational costs and energy consumption. At the same time, sustainable AI practices, such as reducing model complexity to lower energy demands, may limit the interpretability of AI decisions, particularly in high-stakes domains like healthcare. Therefore, one key factor in this tradeoff is the choice between intrinsically interpretable models that consume fewer computational resources, and post-hoc explainability methods that require additional explainability layers. The tradeoff also extends to model training and inference phases. Highly transparent models tend to be less computationally demanding, leading to lower energy usage during both development and deployment. However, in cases where medical AI applications require deep learning approaches, the reliance on resource-intensive post-hoc explainability methods can significantly increase the carbon footprint. 

To address this tradeoff, AI developers should prioritize inherently interpretable models whenever feasible. However, hybrid approaches can also balance sustainability and transparency by reserving complex black-box models and the post-hoc explainability techniques for critical high-risk decisions while using interpretable models for routine or confirmatory analyses. Ultimately, balancing transparency and sustainability requires a multi-stakeholder approach to determine the optimal level of interpretability needed for clinical decision-making while minimizing the environmental impact of AI deployments in healthcare.

\subsection{Sustainability versus Data Governance}\label{subsubsec518}

Sustainable AI development seeks to minimize energy consumption and reduce the environmental impact of data-intensive processes. Such goals may go against traditional governance practices or regulatory compliance that emphasize comprehensive data collection and retention. One of the key aspects of this tradeoff is the use of synthetic data, which offers a sustainable alternative to real-world data collection by promoting the storage, retrieval and use of generative AI models instead of massive raw data \cite{meiser2024survey}. 

In healthcare, gathering real patient data is a time-consuming, resource-intensive, and privacy-sensitive process that requires extensive regulatory oversight. This approach can enhance sustainability by minimizing the need for repeated real-world data collection, reducing the storage burden, and lowering the carbon footprint associated with managing vast medical datasets. However, ensuring the energy efficiency of synthetic data generation is crucial as some algorithms such as diffusion models might consume more energy and resources than the collection of real-world data itself.
To balance this tradeoff, hybrid approaches combining real-world and synthetic data can be beneficial. Selective real-world data collection can be used to validate and refine synthetic data generation models, ensuring that the resulting datasets meet governance standards while minimizing their environmental impact.

\subsection{Accountability versus Transparency}\label{subsubsec519}

The tradeoff between accountability and explainability arises from the tension between the provision of comprehensive, system-wide transparency (global explainability) and the guarantee of case-specific, actionable insights (local explainability) that allow stakeholders (i.e. clinicians, regulators and patients) to assess decision-making processes and assign responsibility \cite{procter_holding_2023}. However, global explainability techniques, which offer a high-level understanding of how an AI model functions across all cases, may not always provide the necessary granularity for accountability in specific clinical decisions. A model may be deemed generally transparent, but still lack sufficient evidence for trustworthiness in an individual diagnosis or treatment recommendation.

Therefore, local explainability methods, which focus on interpreting AI decisions at the case level, are often more effective for clinical accountability, as they allow clinicians to assess why a model produced a particular prediction for a given patient. However, prioritizing local over global explainability might limit a system-wide evaluation of fairness, robustness, and reliability, leading to challenges in regulatory compliance and ethical auditing. To balance accountability and explainability, AI developers and healthcare stakeholders must align transparency methods with the context of use, ensuring that both global system oversight and localized interpretability are available where needed.

\subsection{Accountability versus Privacy}\label{subsubsec5110}

AI accountability requires traceability and auditability, allowing health regulators and stakeholders to assess the model’s decision-making process and distill further responsibilities. However, auditing AI systems often involves accessing sensitive data, model parameters, and training processes, which can conflict with privacy. Full transparency in auditing could inadvertently expose patient information, leading to potential privacy breaches.

To balance accountability and privacy, privacy-preserving auditing techniques must be adopted that enable regulatory bodies to validate AI models without directly accessing patient data. Additionally, model documentation and explainability-by-design approaches can enhance accountability without compromising data confidentiality. However, privacy-preserving strategies may limit the depth of audits, making it harder to detect biases or assess long-term reliability. Therefore, AI developers, regulators, and healthcare providers must collaborate to design auditing frameworks that uphold both transparency and patient privacy, ensuring that AI systems remain trustworthy, ethical, and legally compliant.

\subsection{Human Agency and Oversight versus Technical Robustness and Fairness}\label{subsubsec5111}

A critical trade-off in the implementation of TAI principles in healthcare arises between human agency and oversight and the principles of technical robustness, safety, and fairness. A robustness-driven design of an AI system aims to minimize variability, enforce safety boundaries, and maintain performance under uncertain or adversarial conditions. While these safeguards are vital for ensuring reliable operation, they may inadvertently constrain the autonomy of the clinician, particularly when AI recommendations are treated as default or mandatory in practice. Similarly, fairness-enhancing interventions—such as the suppression of sensitive but potentially informative features or post-processing adjustments—aim to prevent discriminatory outcomes across demographic groups. However, these modifications may reduce the model's capacity to support nuanced, individualized care, potentially limiting the clinician’s ability to tailor decisions to complex patient profiles. In both cases, the clinician's expertise may be diminished by rigid algorithmic constraints, leading to reduced trust, accountability, and flexibility in medical decision-making.

To address this trade-off, AI systems should be designed to promote shared decision-making and adaptive support, rather than rigid automation. Strategies such as graded autonomy \cite{hussain2024development} allow clinicians to override or contextualize AI recommendations, while explainability layers can clarify how robustness or fairness constraints have influenced a given output. Systems might also incorporate clinician-in-the-loop feedback mechanisms that adjust model behavior based on real-time inputs. Ultimately, balancing robustness, fairness, and human agency requires ongoing collaboration between AI developers and healthcare professionals to ensure that ethical standards, clinical expertise, and patient-centered care remain at the core of AI-assisted healthcare.

\subsection{Human Agency and Oversight versus Transparency}\label{subsubsec5112}

While transparency is a cornerstone for fostering trust and enabling informed decision-making, greater transparency (e.g., showing full model logic or uncertainties) can overwhelm clinicians or patients who can experience information fatigue or confusion, especially if explanations are overly technical or misaligned with their decision-making needs. Thus, a clinician may struggle to make timely decisions, potentially defaulting to the model's suggestion—reducing agency despite increased transparency.

To reconcile this trade-off, transparency must be  tailored, and aligned with user capacity and context. Implementing role-specific and layered explanations allows clinicians, patients, and other stakeholders to access the level of detail appropriate to their expertise and needs—ranging from simplified rationales to in-depth statistical outputs. Importantly, co-designing explanation strategies with end-users, especially clinicians, can help preserve their agency without compromising on ethical or regulatory transparency mandates. Ultimately, the goal is to support—not overload—clinical reasoning through intelligible and clinically relevant information.
\section{Challenges towards the practical adoption of the design framework in healthcare}
\label{sec6}

In the health domain AI systems are barely considered and tested on the interaction with clinicians and patients, being still limited to 'proofs of concept' stage and fostering  high accuracy rates that improve objective diagnostic or prognostic metrics. Furthermore, AI systems often struggle with a lack of involvement of end users (clinicians and patients) in defining clinical user requirements, as well as in the involvement of stakeholders in the rest of the life-cycle phases of AI systems \cite{szabo_clinicians_2022}. This leads to a failure in the AI implementation due more to socio-technical factors than pure technical conditions \cite{saw_current_2022}. This framework is aimed at being used in the design phase to facilitate AI developers to meet the principles of TAI in further development but also to make the rest of healthcare stakeholders aware of what they should demand when asking for a TAI medical system. Therefore, with this framework we promote the involvement of the stakeholders from an early phase, to set the basis of the AI system and thus enhance its development and validation, facilitating as well its further clinical adoption. 
 
\paragraph{Algorithm aversion and the need for human oversight}Currently, AI is experiencing an overwhelming boost and people are ecstatic about the potential impact in any life's corner. One aspect, which is seen as a menace for the workforce, not only in healthcare but also in other fields, is that AI will replace workers with an improved performance that surpasses human capabilities. Perhaps the most practical and immediate application of AI in health is in the 'back office' where repetitive tasks that require large volumes of information can be automated such as clinical coding, prior authorization for treatment, managing clinical workflow, scheduling and fraud detection \cite{hashiguchi_fulfilling_2022}. Güngor et al. \cite{gungor2020creating} observed in a survey that although business entities have positive feedback towards AI deployment, society still perceived AI as bringing more risk than value in specific domains, including healthcare. For instance, healthcare professionals are quite hesitant about AI potential benefits, as they do not think that an AI system is capable of bearing the moral responsibility that medical decision making entails. This idea is further highlighted by Maris et al. \cite{maris_ethical_2024} in their study, where some participants expressed that they would be more forgiving of human error compared to AI errors. Possibly this points to a phenomenon commonly referred to as ``algorithm aversion'', where people tend to assess human actions more favorably compared to an algorithm performing the same action. A viewpoint where the human doctor acts as a sparring partner is crucial to interpret the relevance of AI recommendations with the patient, on an equal moral footing, to ensure that practices of shared decision-making occur in an effective, and valuable manner.
 
\paragraph{Human augmentation and the risk of overreliance} AI should not be regarded as a replacement for decision-makers, but as an augmentation to produce accurate algorithmic predictions, which are then certified and overseen with the value and judgment of human experts. Thus, the AI predictive capability must be accounted for as an input in the clinical decision-making process, where the clinical expert's final decision remains critical.  In other words, this `human touch' or `human dimension' is required and preferred by patients in the clinical practice, especially with AI system in the loop, as it ensures compassionate and dignified care that considers not only physical but also the mental well-being of patients \cite{maris_ethical_2024}. This human dimension also embraced human limitations such as biases, fatigue and fallibility, against which participants contrasted certain favorable attributes of AI.
This agency support evolves towards the human augmentation paradigm by which the tuple of human plus AI outperforms both human and AI individually, and implies numerous benefits in the different healthcare processes. The human augmentation in healthcare might transform the term of AI to 'augmented intelligence' reflecting the enhanced capabilities of human clinical decision making when coupled with those computational methods and systems \cite{crigger_trustworthy_2022, drabiak_ai_2023}. Nevertheless, an automation bias might occur due to this overreliance implying a reduction in clinicians' effort to verify AI systems which impacts on the quality, performance, and reliability on the health processes. 

\paragraph{On the accountability of AI-based clinical decision making}Despite the positive support that AI systems provide to clinicians in the decision making as well as the clinicians' positive attitude towards using AI, they manifest that at the end of the day it is the physician who must take responsibility and make the final decision about a diagnosis, based on a value judgment which AI lacks as it is not capable of bearing the moral responsibility that medical decision-making entails \cite{gondocs_ai_2024, maris_ethical_2024}. One of the most relevant challenges to turn the AI into a reality in the healthcare domain is to elaborate on the accountability of the medical decision when using AI assisted systems. Proper accountability measures should be designed based on the application domain. For instance, in medical AI, users of the system, such as doctors, can be held accountable for the harm caused by the system because they are domain experts in the field. Therefore, they should only use AI systems to assist their decision-making, not for making decisions for them \cite{kaur_trustworthy_2022}. In the event of a wrong prediction, the boundaries of accountability are blurred as it is difficult to precise where the error happened in the chain of the decision making. We could tend to think that the responsibility of the stakeholders towards the event when a wrong prediction is made relies mainly on the stakeholders that interact with the model or are affected by its prediction, as they are the ultimate enforcers of the model's decision. This could retract stakeholders from using AI systems due to the consequences that an erroneous decision by the model could take. Thus, here it is essential to propose regulatory actions to standardize processes that drive the accountability of the models and their stakeholders that promote a set of actions that facilitate the audit of the potential problems such as an extensive report of the model that cover not only the algorithmic approaches but also other elemental points as training data, etc. In this case, a regulatory framework as well as a regulatory agency, or even the legal services from HC providers, that oversees this principle would be beneficial to minimize issues that arise when a wrong prediction occurs \cite{wiegand2019and}. However, regardless of the role of the different stakeholders and their commitment to a potential regulatory framework, AI systems will need to be capable of being accountable for their decisions in ways that clinicians can trust \cite{procter_holding_2023}.

\paragraph{Continuous involvement of all stakeholders throughout the AI lifecycle}The goal of this TAI framework is to facilitate developers in designing and building more trustworthy medical AI systems, while also equipping healthcare stakeholders with a structured set of requirements they can demand when procuring or deploying AI tools to support clinical processes. By aligning design practices with the principles of TAI, the framework helps ease compliance with regulatory frameworks that mandate ethical, transparent, and safe AI, such as the EU Artificial Intelligence Act. In particular, the AI Act classifies AI systems used in medical diagnosis, monitoring, or treatment as high-risk, requiring conformity assessments, documented risk management strategies, and mechanisms to ensure transparency, human oversight, and data governance. Our framework addresses these dimensions early in the development cycle, offering guidance that supports developers and healthcare institutions in fulfilling regulatory obligations while fostering safe and ethically aligned adoption of AI in healthcare. However, while the framework provides a valuable structure to promote TAI-compliant systems, its effective use will depend on continuous engagement of AI developers with regulators, multidisciplinary teams, and clinical end-users.

\paragraph{Interpretability of AI explanations}In the field of AI applied to health, clinicians' trust is struggling, as it is more reliant on their experience than on the AI solution. As we have seen, one of the TAI principles with the highest impact on users' trust is explainability, which adds an additional layer of complexity to the AI solution. Unfortunately, too much explanatory information can be overwhelming and not always easy to interpret, making it not useful in the clinical context. Thus, exploring simpler explanations for users with non-technical backgrounds is key to improving the adoption of AI solutions. In this case, however, a tradeoff between the usability of the model and its transparency could arise \cite{kacafirkova_trustworthy_2023}. Even when explainability is recognized to be user-dependent evaluating, the clarity of explanation, even with example-based methodologies is not straightforward. In some cases, such as feature visualization the interpretation of the saliency maps is still subjective and requires careful consideration to ensure meaningful insights \cite{nasarian_designing_2024}. The assessment of the explanation could be reinforced as they are provided with a ground truth that revolves around the causality of the decision, eventually making the evaluation and quantification of the quality of the explanation more reachable \cite{salahuddin_transparency_2022}. 

\paragraph{Misalignment of explanations issued for model surrogates}Another limitation is that unfaithful explanations can lead to misinformed decisions, particularly when using surrogate models—interpretable models that approximate the behavior of more complex black-box systems. These are often employed in the medical field when the original models are inaccessible due to security or implementation constraints. However, if the surrogate provides the same prediction as the original model but for a different rationale, clinicians may be misled into focusing on the wrong symptoms or diagnostic tests, potentially resulting in inappropriate treatment or mismanagement of the patient’s condition \cite{mariotti_beyond_2023}. To mitigate this, linking explanations from surrogate models with ground truth can help detect and reduce misleading explanations.
 
\paragraph{From static AI-based clinical systems to interactive multi-agent committees}Another element to guarantee trust in the explanations is their interactivity. Now that LLMs are bursting into the AI domain with great impact, incorporating natural language dialogues in the explanations is of particular interest, so clinicians can deal with the AI solutions as if they were ``professional colleagues''.  A forward-looking scenario envisions multi-agentic systems in which several AI agents, each emulating different clinical reasoning styles, collaborate with one another and interact directly with human clinicians to discuss specific patient cases. This setup would resemble a virtual medical committee, enhancing the interpretability and contextual grounding of the AI's outputs. To address the interactivity of explanations, the human-AI interface plays a fundamental role if it contains two types of elements: (1) information (on request) about the scientific validation and evaluation of the performance of the solution, such as the reporting data used to train and test the models, (2) medically (not technically) relevant factors in the decision-making of the AI solution and how they are connected to the logic of the algorithm \cite{mueller_explainability_2022}.

\paragraph{Measuring trust on explanations}Despite the potential benefits brought by explainable AI, one of the grand challenges to be addressed when used in the medical practice is how to evaluate the relevance of the explanations, or the suitability of subjective scales aimed to measure trust on the issued explanations \cite{hamon_bridging_2022, kacafirkova_trustworthy_2023}. Metrics not only build trust in healthcare, but also promotes the acceptance of AI assistance among practitioners. By having access to quantifiable metrics that assess the significance and relevance of the XAI explanations, practitioners can gain confidence in the decisions made by AI systems. This, in turn, contributes to the establishment of a critical trust in AI technologies, where practitioners can rely on the assistance provided while maintaining their professional judgment, important as erroneous or misleading decisions can be significant. 

\paragraph{Informative explanations for causal relationships discovered by medical AI systems}An evaluation strategy of XAI methods should capture trust, causality, and informativeness \cite{Nauta2023_csur_evaluating-xai-survey, pietila_when_2024}. Trust can be defined as a combination of various aspects: trust in the model performance, trust in the model accuracy, subjective perception by the user, and the similarity between model behavior and human behavior in making decisions (i.e., the model makes mistakes in the same cases of human mistakes). Causality is related to the capability of inferring properties of the natural world by AI models. In medicine, this is a crucial aspect of AI models because it makes it possible to reveal a strong association between features and events. Likewise, an explanation should be informative, and human decision-makers should be provided with information that can be useful for them. Informativeness concerns the capability of explanation models to provide valuable information regarding the decision process. In the field of medicine, it is crucial to have XAI methods that increase trust by explaining the causal relationships discovered by medical AI systems \cite{scarpato2024evaluating}. However, the benefits of such explainability risk being undermined if the evaluation metrics used to assess these explanations are themselves overly complex or opaque. In any case, XAI evaluation should not replace one challenge—evaluating explanations—with another—explaining the metrics. Especially in clinical settings, where many stakeholders may not have formal training in computer science, statistics, or mathematics, evaluation metrics must remain simple, clear, and interpretable to ensure meaningful validation and foster genuine trust.

\section{Conclusions and outlook}
\label{sec7}

The rapid integration of AI technologies into the healthcare sector presents unprecedented opportunities to enhance clinical decision-making, improve diagnostic accuracy, and optimize patient care. However, these benefits are accompanied by significant ethical, legal, and societal challenges, particularly concerning the trustworthiness of AI systems. AI application in medicine must not only demonstrate high performance for the global good, but also uphold ethical aspects that protect patient rights and maintain professional integrity. Despite growing attention to TAI, there remains a lack of concrete, actionable frameworks tailored to the healthcare domain that account for the needs of diverse stakeholders. Thus, ensuring that medical AI systems adhere to TAI principles is essential for their safe and effective integration into healthcare. 

This paper is motivated by the need to bridge that gap by proposing a design-phase framework that operationalizes TAI principles in medical AI systems. The proposed framework is intended to serve a dual purpose: 1) guiding AI developers in embedding trustworthiness from the beginning of the system design phase, and 2) offering healthcare stakeholders a structured foundation to evaluate and demand essential trust-related features. A key contribution of this work is its emphasis on how TAI principles manifest in the diverse interactions between healthcare stakeholders and AI systems across various clinical processes. For each TAI subprinciple, we propose a set of actionable requirements, considering both the degree of stakeholder involvement and the prioritization of these requirements through a MoSCoW-based ranking scheme. Furthermore, we move beyond abstract recommendations by demonstrating how the proposed framework can be applied in practice, using the AI-assisted diagnosis and prognosis of cardiovascular diseases (CVD) as a use case. In doing so, this framework aims to support the responsible and effective integration of AI into clinical workflows, ultimately promoting transparency, fairness, safety, and accountability in medical decision-making.

With the aim of serving as a comprehensive reference for two key audiences (AI developers and healthcare stakeholders), this paper has first outlined the growing benefits and pressing need for integrating TAI into medical AI applications. We have established the conceptual foundations of TAI, presenting its core principles alongside relevant international standards that align with these values. From a healthcare perspective, we have examined the primary medical processes, data types, and stakeholders involved, illustrating how they interact with AI systems at different stages of the care continuum. The central contribution of the paper is structured into three core components. First, we have described the design framework, translating TAI principles into concrete system requirements, supported by existing works and current trends within each TAI principle's domain. We have further outlined methodologies to evaluate the fulfillment of these requirements and demonstrate their application through the CVD use case. Next, we have explored several key tradeoffs that can emerge between competing TAI principles, proposing strategies to ameliorate these tensions in medical contexts. Finally, we have discussed broader challenges posed by the integration of AI into healthcare, including concerns about AI surpassing human performance, the evolving role of humans in augmented decision-making, limitations in explainability, and the critical need for explanation validation. While not exhaustive, our discussion in this regard has exposed the underlying tradeoffs in achieving TAI, underscoring the relevance of the proposed framework to address them.

To guide the development and assessment of the proposed framework, we have formulated four RQs focused on identifying, contextualizing, and assessing the requirements that a medical AI system must meet to be considered trustworthy from a healthcare perspective. These questions serve to operationalize the TAI principles by exploring their concrete implications across medical processes and stakeholder interactions. The structure of the paper has been aligned with these RQs, offering both theoretical grounding and practical considerations to reply them as follows: 
\begin{itemize}[leftmargin=*]
\item \textit{RQ1: What requirements must a medical AI system meet to comply with TAI principles?} 

To address this first question, the paper has presented a comprehensive set of requirements derived from a systematic analysis of the TAI principles—namely, (1) Human agency and oversight, (2) Technical robustness and safety, (3) Privacy and data governance, (4) Transparency, (5) Diversity, non-discrimination and fairness, (6) Societal and environmental well-being (sustainability), and (7) Accountability. Each principle has been further broken down into relevant sub-principles to allow for a more granular understanding of how trustworthiness could be addressed in medical contexts. For each sub-principle, we have reviewed and synthesized key areas of concern and related work, translating these insights into specific, actionable requirements that medical AI systems should fulfill to ensure alignment with the respective trust criteria. To support implementation efforts, we have also proposed a qualitative prioritization scheme to guide stakeholders in identifying the most critical requirements. Each requirement is assigned one of three priority levels based on its essentiality for aligning with TAI principles. This prioritization offers a pragmatic lens through which both AI developers and healthcare stakeholders can assess which requirements are indispensable, which can be recommended, and which can enhance trust alignment when feasible. This layered structure ensures that the requirements are not only ethically grounded, but also adaptable to the practical constraints and specificities of different healthcare contexts.

\item \textit{RQ2: Would healthcare stakeholders interact differently with these features, and if so, how?} 

To address RQ2, we have first considered the diverse set of stakeholders involved in medical processes where AI systems are deployed—namely, patients, clinicians, healthcare providers, and healthcare regulators or policymakers. Each of these stakeholders interacts with or is affected by medical AI systems in distinct ways, depending on their roles, responsibilities, and proximity to the system’s decisions. As such, their demands and expectations regarding the trustworthiness of AI solutions naturally diverge. Recognizing these differences, we have reflected these heterogeneous interactions in a matrix requirements-stakeholders roles for each TAI principle. This matrix includes a range of stakeholder roles (Responsible, Affected, Uses, Monitors, Contributes, and Informed) that map to different levels of engagement within the medical AI lifecycle. For example, clinicians are primarily concerned with agency and oversight, patients prioritize privacy, explainability, and robustness, and healthcare providers and regulators focus more on fairness, accountability, and system-level safety. By delineating the roles and interactions, we ensure that the proposed framework accommodates the complex, multi-actor landscape of medical AI. Additionally, in the CVD diagnosis and prognosis use case, we have illustrated how clinicians and patients have distinct needs regarding AI systems, and how the proposed requirements can be articulated to effectively address those differences. 

\item \textit{RQ3: How can the fulfillment of these requirements be measured, and what key aspects should be assessed to ensure compliance?} 

Despite the growing emphasis on TAI, the rigorous evaluation of TAI principles remains an underexplored area. This gap is particularly critical in the healthcare domain, where the assurance of addressing these principles must be supported by systematic and verifiable assessment methods. Ensuring that an AI system aligns with TAI principles is not only a matter of good design practice, but also requires methodologically sound evaluation frameworks that can determine whether the system truly meets the expectations of safety, fairness, transparency, and other core principles. In our work, for each requirement proposed under the TAI principles, we have pointed to existing literature and emerging trends that suggest possible directions for evaluation. While an in-depth development of specific evaluation frameworks goes beyond the scope of this work, we emphasize the importance of prioritizing this area in future research. We have aimed to raise awareness about the kinds of qualitative and quantitative approaches (ranging from risk assessments and stakeholder interviews to performance metrics and audit trails) that could form the basis of such validation. These approaches would be instrumental to determining to what extent a medical AI system is aligned with the TAI framework, offering the foundation for formal certification or trust labels in future regulatory ecosystems.

\item \textit{RQ4: What strategies could be employed to manage the tradeoffs arising between different TAI principles?}

In response to this last question, it is important to note that the framework proposed in this paper is conceived as a comprehensive guide for addressing and implementing requirements in the design and development of medical AI systems that aligns with TAI principles. However, it is not intended to be a \emph{one-size-fits-all} solution. The relevance and implementation of each requirement will necessarily vary depending on the specific context, clinical setting, and intended function of the AI system. Given this variability, tensions and tradeoffs will surely emerge between different TAI principles, reflecting the complex interplay between ethical imperatives, technical constraints, and stakeholder priorities. In the article we have identified and examined a series of tradeoffs that frequently occur in the healthcare domain, such as accuracy versus fairness, privacy versus transparency, or fairness versus data governance. For each of these, we have proposed potential strategies to mitigate their impact while maintaining alignment with the core goals of TAI. Nonetheless, we acknowledge that this list is not exhaustive. Tradeoffs may also arise from the specific design, deployment context, or target population of a particular medical AI tool. Therefore, careful reflection and iterative dialogue among AI developers, clinicians, patients, and regulatory bodies remain essential to balance these tradeoffs effectively and responsibly.
\end{itemize}

Future work should address several important extensions to the framework proposed in this study. First, the rapid emergence and widespread adoption of Generative AI (GenAI) technologies (including Large Language Models and Multimodal  Diffusion-based architectures) have begun to reshape the landscape of medical AI applications. While the current framework is designed to be agnostic to specific AI models or techniques, it is clear that GenAI introduces new avenues for future research in terms of explainability, data provenance, misuse potential, and even authorship of medical content. These challenges call for a refined perspective on how TAI principles can be meaningfully applied to GenAI systems, especially as their use expands into clinical decision support, synthetic medical data generation, and patient communication tools. Future research should therefore explore how TAI principles—such as accountability, robustness, and transparency—can be adapted and operationalized in GenAI-based solutions.

Secondly, although the focus of this paper has been on the design phase of medical AI systems, the need to evaluate whether the proposed requirements are being met is essential to closing the loop between design intent and real-world performance. As discussed in RQ3, the validation and assessment of the fulfillment of the TAI requirements is not only a scientific challenge but a regulatory one, particularly in light of the AI Act. Therefore, AI systems deployed in the healthcare domain will be subject to rigorous conformity assessments that demonstrate alignment with TAI principles across their lifecycle. Therefore, a key research direction to be pursued in the future will be to establish comprehensive, principle-specific qualitative and quantitative evaluation frameworks that allow for the traceable verification of trustworthiness claims. Such frameworks will be essential to give stakeholders means to assess whether AI tools respect patient rights, support clinicians’ decision-making, and meet ethical and legal standards. 

Lastly, this work has focused on the healthcare domain with examples for CVD, where AI has perhaps the greatest potential for the social good. However, the underlying methodology and design rationale of the proposed framework are transferable to other domains. With appropriate domain-specific adjustments, the TAI-by-design approach proposed here could serve as a foundation for trust-aware AI development in other sensitive and high-stakes fields such as education, law, social services, or finance. Each of these sectors involves its own unique stakeholder interactions, regulatory environments, and societal risks. Extending the framework to these contexts would involve rethinking requirements, priorities, and tradeoffs, while maintaining their shared central goal: to ensure that AI systems are not only technically capable, but also aligned with human values and institutional responsibilities.

On a closing note, this paper has presented a design-phase framework for operationalizing TAI principles in the context of medical AI systems. We envision this framework as both a practical reference for AI developers and a foundation for healthcare stakeholders to actively demand and co-develop AI solutions that meaningfully support clinical practice while prioritizing patient safety and well-being. Looking forward, we encourage the research community, healthcare professionals, and industry partners to adopt, adapt, and expand this framework through interdisciplinary collaboration and empirical validation. Advancing this collective effort in real-world settings is essential to fostering the responsible, transparent, and sustainable integration of AI tools that truly empower clinical decision-making and improve healthcare outcomes.

\section*{Acknowledgments}

P. A. Moreno-Sanchez and J. Del Ser thank the support of the Basque Government through EMAITEK and ELKARTEK funding grants (KK-2024/00064, KK-2024/00090) and the consolidated research group MATHMODE (IT1456-22). P. A. Moreno-Sanchez and M. van Gils received support from the project PerCard (Personalised Prognostics and Diagnostics for Improved Decision Support in Cardiovascular Diseases) in ERA PerMed via the Research Council of Finland (decision number 351846), under the frame of ERA PerMed.

\section*{Declaration of generative AI and AI-assisted technologies in the writing process}

During the preparation of this work the authors used GPT4o/ChatGPT in order to improve the readability and language of the manuscript. After using this tool/service, the authors reviewed and edited the content as needed and take full responsibility for the content of the published article.








\bibliographystyle{elsarticle-num} 
\bibliography{TAI-bydesign-Health.bib}

\end{document}